\documentclass[journal]{IEEEtran}

\usepackage{graphicx}
\usepackage{xcolor}
\usepackage{amsmath}
\usepackage{amssymb}
\usepackage{mathtools}
\usepackage{amsthm}
\usepackage{tikz,lipsum,lmodern}
\usepackage[most]{tcolorbox}
\usepackage{subcaption}
\usepackage[nocompress]{cite}

\usepackage{hyperref}

\newcommand{\tomt}[1]{\textcolor{black}{#1}}

\title{Deep Internal Learning: \\ Deep Learning from a Single Input}
\author{Tom Tirer~\IEEEmembership{Member,~IEEE}, Raja Giryes~\IEEEmembership{Senior Member,~IEEE}, Se Young Chun~\IEEEmembership{Member,~IEEE} and Yonina C. Eldar~\IEEEmembership{Fellow,~IEEE}
\thanks{Tom Tirer is with the Faculty of Engineering, Bar-Ilan University, Ramat Gan 5290002, Israel (e-mail: tirer.tom@gmail.com). Raja Giryes is with the School of Electrical Engineering, Tel Aviv University, Tel Aviv 69978, Israel (e-mail: raja@tauex.
tau.ac.il). Se Young Chun is with the Department of Electrical and Computer Engineering, Seoul National University, 1 Gwanak-ro, Gwanak-gu, Seoul 08826, Korea (e-mail: sychun@snu.ac.kr). Yonina C.~Eldar is with the Faculty of Mathematics and Computer Science, Weizmann Institute of Science, Rehovot 7610001, Israel (e-mail: yonina.eldar@weizmann.ac.il).}}
\date{February 2022}

\begin{document}

\maketitle

\begin{abstract}
    Deep learning, in general, focuses on training a neural network from large labeled datasets. Yet, in many cases there is value in training a network just from the input at hand. 
    This is particularly relevant in many signal and image processing problems where training data is scarce and diversity is large on the one hand, and on the other, there is a lot of structure in the data that can be exploited. Using this information is the key to \emph{deep internal-learning} strategies, which may involve training a network from scratch using a single input or adapting an already trained network to a provided input example at inference time. This survey paper aims at covering deep internal-learning techniques that have been proposed in the past few years for these two important directions. 
    While our main focus will be on image processing problems, most of the approaches that we survey are derived for general signals (vectors with recurring patterns that can be distinguished from noise) and are therefore applicable to other modalities.
\end{abstract}

\section{Introduction}

Deep learning  
\tomt{methods have}
led to remarkable advances with excellent performance in various fields including natural language processing, 
optics, 
image processing, 
autonomous driving, 
text-to-speech, 
text-to-image, 
face recognition, 
anomaly detection, 
and many more applications.
Common to all the above advances is the use of a deep neural network (DNN) that is trained using a large annotated dataset that is created for the problem at hand. The used dataset is required to represent faithfully the data distribution in the target task, and allow the DNN to generalize well to new unseen examples.
Yet, achieving such data can be burdensome and costly and having strategies that do not need training data or can easily adapt to their input test data is of great value. This is particularly true in applications where generalization is a major concern such as clinical applications and autonomous driving.

In scenarios, where no training examples are available for a given problem, or one does not want to learn from examples that may not faithfully represent the true data, one is required to train the DNN only on the given input example. This may involve exploiting prior knowledge of the problem such as internal self-similarity between patches in an image~\cite{ulyanov2018deep,shocher2018zero,shaham2019singan} or exploiting common models in signal processing such as sparsity and other regularizers. Even in the case where training data do exist, training or fine-tuning a DNN on the input image can be useful in order to better adapt the DNN to its statistics. The input image may not be well represented in the training data~\cite{tirer2019super,IAGAN} and therefore the input image can be used as another source for training the network, to improve its performance. Structure on the data in the form of regularizers can also compensate for missing data, and enables the use of many well developed signal processing tools and concepts.

In this survey paper, we aim at covering \emph{internal-learning} techniques that allow training DNNs on a given input example. We will see here how the use of signal processing elements such as models, statistics, priors and more can be utilized to compensate for the lack of data, forming a bridge between traditional signal processing tools and modern deep learning.
We divide our discussion into techniques that only train on the input example \cite{ulyanov2018deep,DoubleDIP,Cheng2019Bayesian,DeepDecoder,liu2019image,Jo_2021_ICCV,Zukerman20BPDIP,Sun21PnP,mataev2019deepred,Metzler:2019gsure,Abu-Hussein_2022_WACV,shocher2018zero,Quan20Self2Self,shaham2019singan,nikankin2022sinfusion,kulikov2023sinddm}
and approaches that use a pre-trained network but fine-tune on the input example at test time \cite{
soltanayev2018training,Zhussip_2019_CVPR,zhussip2019extending,kim2020unsupervised,soh2020meta,park2020fast,IAGAN,tirer2019super,pan2020dgp,mohan2021adaptive,abu2022adir} for the tasks of reconstruction or generation / editing.
Diverse internal learning strategies will be surveyed such as self similarity, multiscale similarity, meta learning, statistical loss functions, consistency loss functions, and above all the use of a network structure as a prior.

Our focus in this paper will be on internal learning in the context of deep learning. Note that there is rich literature about internal learning in the `pre-deep-learning' era, which we do not cover in this survey. The interested reader may refer to  \cite{irani2019blind}.
\tomt{In addition, we will present mainly deep internal learning approaches in the context of images, which is also the focus of most of the deep internal-learning works.} 
There are also recent efforts to apply internal learning in other modalities, such as audio \cite{greshler2021catchawaveform} and 3D graphics \cite{williams2019deep,Hanocka2020Point2Mesh,Hertz2020Deep,Metzer21Orienting,Metzer21SelfSample}. 
Thus, we believe that many of the ideas surveyed here can  be beneficial in many other signal processing applications.  Exposing the signal processing community to these techniques in a unified manner can aid in promoting further research and applications of these important methods.

\section{Brief background on external learning}

To put our review of deep internal learning in context, let us begin with a brief discussion and formulation of the common theme in machine learning: training DNNs using massive external data in an offline phase. We name this approach ``external learning" to emphasize its distinction from internal learning.

Deep learning methods are often based on massive training sets:
pairs of input samples $\{ x_1,\ldots,x_N\}$ and their corresponding annotations $\{ y_1,\ldots,y_N\}$. 
A DNN architecture $h(\cdot; \theta)$ is then designed and its parameters $\theta$ are optimized in an offline training phase by minimizing a loss function 
\begin{equation}
\tilde{\theta} = \arg \min_\theta \sum_{i=1}^N L(h(x_i; \theta),y_i)
\label{eq:suptrain_}
\end{equation}
such that for a new input $x_0$, the output of the trained DNN $h(x_0; \tilde{\theta})$ approximates the unknown corresponding annotation $y_0$.

For example, in imaging tasks (e.g., low-level computer vision), each $x_i$ in the training set may be a degraded version of an associated ground truth image that is used as the target of the network $y_i=x_{gt,i}$.
Specifically, the common practice to train a DNN for a specific task, e.g., super-resolution with certain downsampling model $f(\cdot)$ \cite{dong2014learning}, is to take a collection of ground truth high-resolution images $x_{gt,1}, \dots, x_{gt,N}$ and generate the low-resolution input samples $x_1=f(x_{gt,1}), \dots, x_N=f(x_{gt,N})$ using the predefined $f$.

\tomt{
{\em Limitations of external learning.}} 
DNNs that are trained using the common 
external learning approach typically perform very well when the assumptions that have been made in the training phase (such as the observation model) are satisfied also by the data at test time. 
However, whenever there is a mismatch between the test and training data, these networks exhibit significant performance degradation \cite{shocher2018zero,tirer2019super}.
Furthermore, oftentimes the degradation model is not known in advance and thus a supervised training approach cannot be utilized.
Another challenge is when ground truth data is scarce or possibly not available.
\tomt{These limitations are inherently bypassed by internal learning:} training a DNN to recover the unknown image $x_{gt}$ using only the test-time observation $x_{0}$.

\section{Overview of internal learning}
\label{sec:overview}

\subsection{What makes internal learning work?} 
\label{sec:overview_core}

There are two complementary factors that are necessary for making internal learning beneficial: the first is information-related and the second is algorithmic-related. 

The information-related condition solely depends on the single observed signal $x_0$: 
{\em recurrence of patterns} or --- using common terminology --- {\em self-similarity}.
Such recurrence, which can be both within and across scales of resolution, allows a suitable learning algorithm to distinguish between components of the signal and random noise or infrequent artifacts. 
Real-world signals, such as images, possess recurring patterns; see e.g. \cite{glasner2009super, zontak2011internal}.

\begin{figure}
    \centering
  \includegraphics[width=250pt]{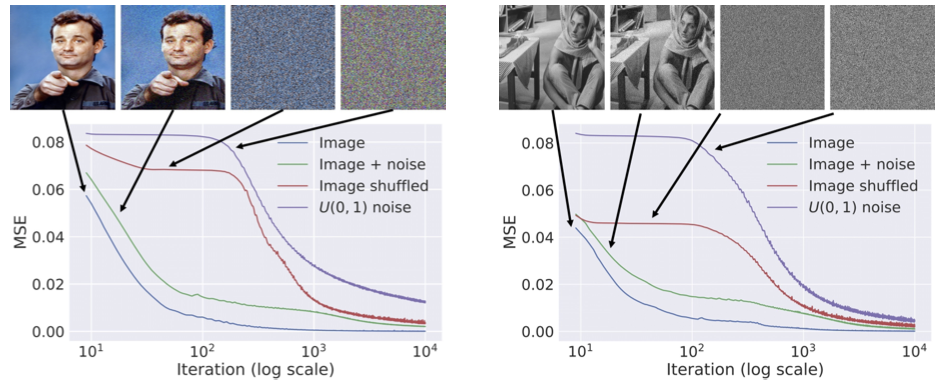} 
  \caption{Learning curves for the reconstruction task using: a natural image, the same plus i.i.d.~noise, the same randomly scrambled, and white noise. Natural-looking images result in much faster convergence, whereas noise is rejected. Figure taken from \cite{ulyanov2018deep} \tomt{and used by permission of the authors}.}
\label{fig:dip_signal_before_noise}
\end{figure}

The algorithmic-related requirement is that the learning algorithm will indeed capture the components of the signal rather than the noise/artifacts even though no explicit supervision is provided and both are `mixed' in the single input sample that is given.
At first glance, this task seems very complicated. Indeed, before the groundbreaking Deep Image Prior (DIP) paper \cite{ulyanov2018deep} was published, it was not clear that modern DNNs, which are highly over-parameterized and can easily (over-)fit the entire noisy sample, would isolate the signal from the noise and artifacts.
Nevertheless, an intriguing experiment from \cite{ulyanov2018deep}, which is presented here in Fig.~\ref{fig:dip_signal_before_noise}, shows 
that the algorithmic requirement is possessed by optimization of a suitable DNN model $x=h(z; \theta)$, with random input $z$, to fit $x_0$, a noisy or pixel-shuffled version of a true clean image $x_{gt}$.
Specifically, Fig.~\ref{fig:dip_signal_before_noise} shows that when optimizing the loss
\begin{equation}
\label{eq:DIP_poc}
\min_\theta \| h(z; \theta) - x_0 \|^2
\end{equation}
with gradient-based methods (e.g., Adam \cite{kingma2014adam}), the DNN fits the clean signal $x_{gt}$ before it fits noise or other pattern-less artifacts. 
Thus, even when $x_0$ is degraded, $x=h(z; \theta)$ estimates the clean signal if the optimization procedure is terminated ``on time" (we elaborate on this point below).

The authors of \cite{ulyanov2018deep} related this behavior to an implicit deep prior that is imposed by the DNN convolutional architecture itself. In \cite{shabtay2023pip} similar behavior was related to positional encoding and implicit representations. More recent theoretical studies on gradient descent and its stochastic variants hint that such simple optimizers have implicit bias (``prior") on their own: a tendency to converge to simple solutions, e.g., with low norms or repetitions, among the many possible solutions that can be realized by an over-parameterized DNN.

\subsection{Brief background on internal learning}
\label{sec:overview_poc}

The proof-of-concept experiment that is presented in Fig.~\ref{fig:dip_signal_before_noise} motivates the general restoration approach that is proposed in the DIP paper \cite{ulyanov2018deep}. There, the observation model is given by
\begin{align}
\label{eq:observation_model}
    x_0 = f(x_{gt}) + e,
\end{align}
where $x_{gt} \in \mathbb{R}^n$ is an unknown true image, $f: \mathbb{R}^n \to \mathbb{R}^m$ is a known forward model/operator and $e \in \mathbb{R}^m$ is the unknown noise (typically assumed to be white and Gaussian).
Focusing on imaging, note that many acquisition processes can be modeled with a linear $f$. For example, blurring (in the deblurring task), downsampling (in the super-resolution task),
and of course, the identity operator $I$ (in the denoising task), are associated with linear instances of $f$.

Let $x=h(z; \theta) \in \mathbb{R}^n$ be a ``hourglass” architecture (also known as “encoder-decoder”, and similar to U-net) and $z$ be a random Gaussian input.
It is proposed to optimize the DNN's parameters $\theta$ by minimizing the least squares loss
\begin{equation}
\label{eq:DIP_loss}
\min_\theta \| f (h(z; \theta)) - x_0 \|^2
\end{equation}
using gradient descent or Adam.
The optimization is terminated via a suitable early stopping (in \cite{ulyanov2018deep}, a maximal number of iterations is manually tuned per task).
Then, the unknown $x_{gt}$ is estimated by $x = h(z; \hat{\theta})$, where $\hat{\theta}$ denotes the DNN's parameters at the early stopping point (i.e., not necessarily a global minimizer of \eqref{eq:DIP_loss}).

The DIP approach excludes any offline training, which typically requires a predefined forward (observation) model $f$ and a collection of ground truth clean training samples. 
Therefore, its main advantage is that it
offers full flexibility to the forward model and data distribution, and avoids the significant performance degradation that is observed when applying an offline trained DNN to a test image whose acquisition mismatches the assumptions that are made in the training phase.

On the other hand, DIP has several major limitations, such as a large inference run-time (since the DNN parameters are optimized in test-time), the need for accurate early stopping to avoid fitting the measurements' noise/artifacts, and the potential performance drop due to not exploiting any data other than the test-time input.
Accordingly, many follow-up works have proposed techniques for addressing these limitations, while keep exploiting the benefits of internal learning.  
This, however, oftentimes requires focusing on a more specific observation model than the general one in \eqref{eq:observation_model} (e.g., certain classes of forward models, $f$, and certain distributions of the noise $e$).

In Box 1 we present several visual results from the application of DIP. Notice the flexibility of this method in terms of the observation model.

\begin{tcolorbox}[breakable,colback=blue!5!white,colframe=blue!75!black,title=Box 1: Internal Learning by Deep Image Prior]

  The Deep Image Prior (DIP) approach \cite{ulyanov2018deep} provides a flexible method to estimate an image $x_{gt}$ from its observations $x_0=f(x_{gt})+e$, where $f$ is a known degradation model and $e$ is noise. In DIP, the estimate is parameterized by a U-net DNN, $x=h(z;\theta)$, with a random noise input $z$ and parameters $\theta$ that are obtained by
  $$
\min_\theta \| f (h(z; \theta)) - x_0 \|^2.
  $$
No offline training phase, based on external data, and no explicit prior terms are used.

The same approach, potentially with some hyperparameter tuning, can be applied to a wide variety of tasks, as shown in the figures below.

\vspace{3mm}

\includegraphics[width=0.95\linewidth]{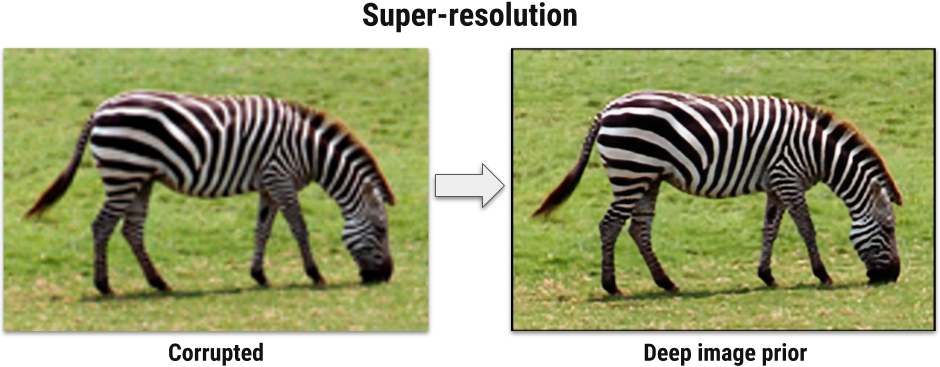}
\includegraphics[width=0.95\linewidth]{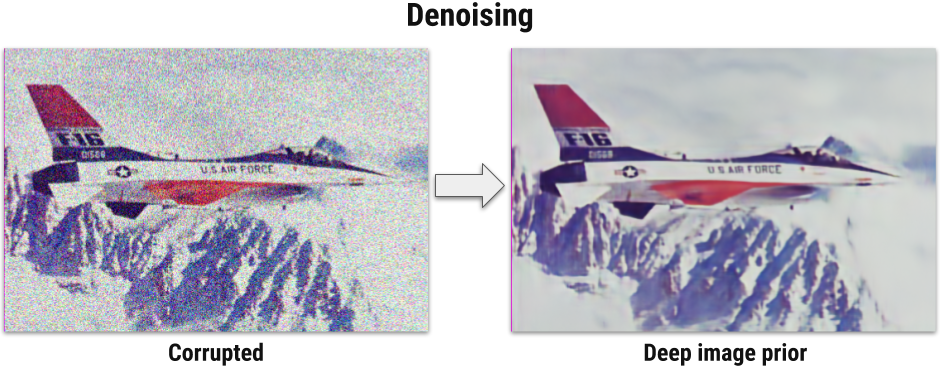}
\includegraphics[width=0.95\linewidth]{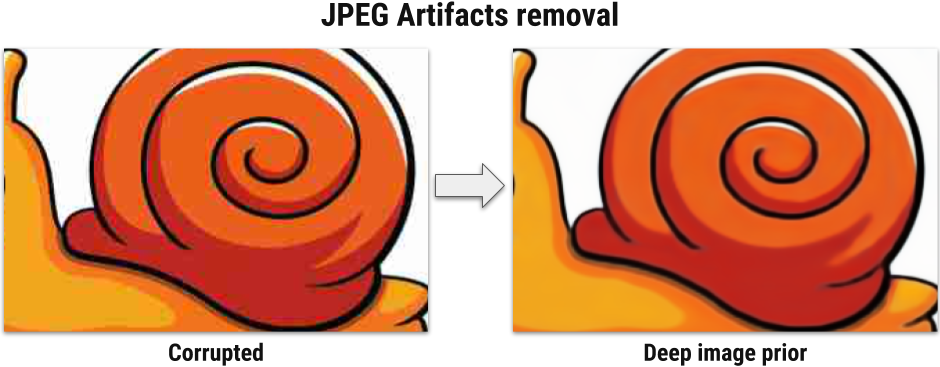}
\includegraphics[width=0.95\linewidth]{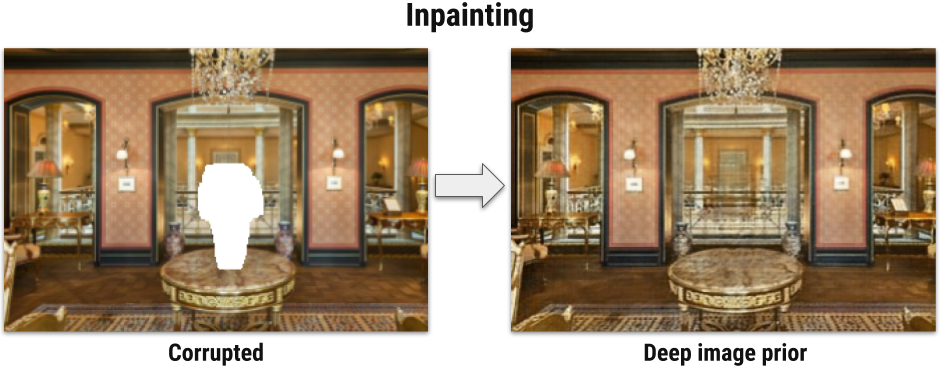}

Figures taken from \cite{ulyanov2018deep} \tomt{and used by permission of the authors}.

The main limitation of DIP is that whenever their observations contain noise or artifacts (e.g., in denoising and JPEG artifacts removal), accurate early stopping is required to avoid fitting them.

\end{tcolorbox}

\begin{figure}
\centering
\includegraphics[width=0.5\textwidth]{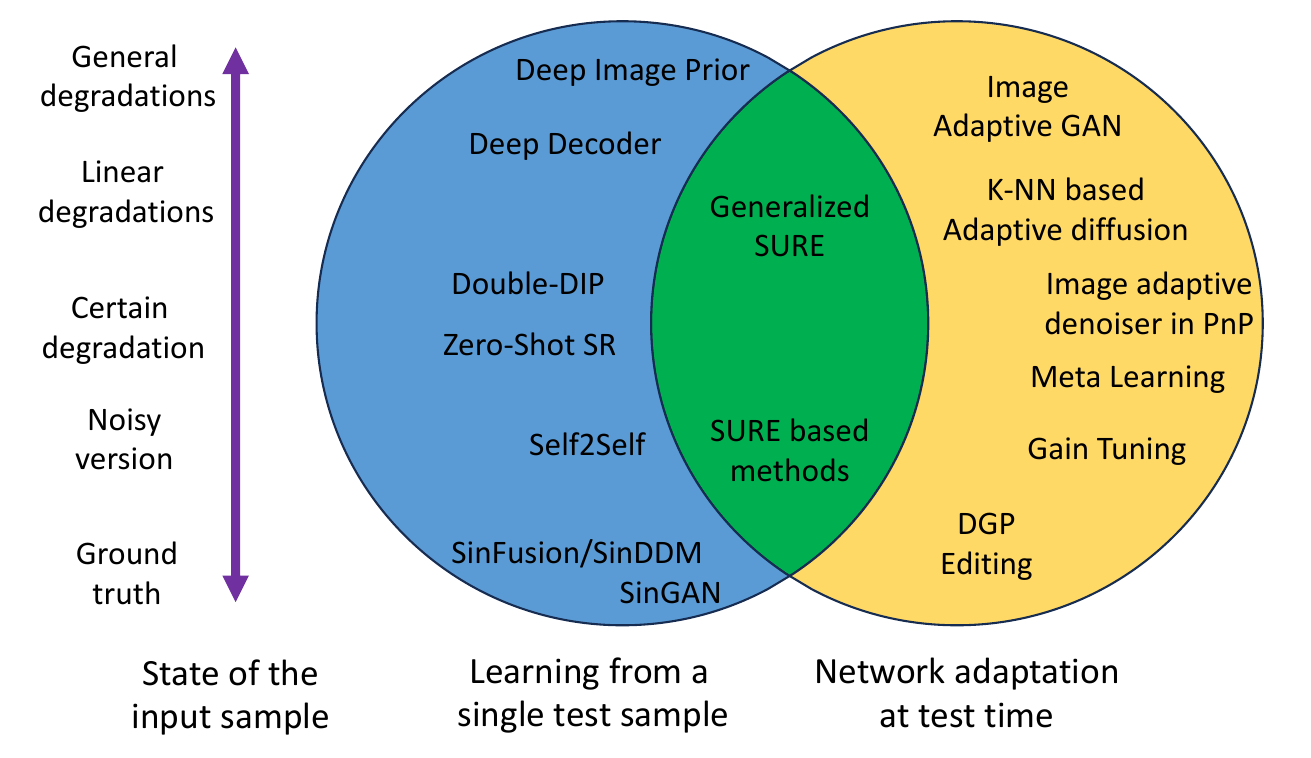}
 \caption{Internal learning approaches can be divided into two high level classes: 1) techniques that learn only from a single example, and 2) techniques that take an already trained network and fine-tune it at test time. Several representative methods are presented, some of which can be utilized for both approaches.
 The vertical axis presents the state of the input sample at test-time ($x_0$ in the paper's notation). Editing/generation techniques require it to be a ground truth (``clean") sample, while reconstruction techniques 
 attempt to recover the unknown ground truth sample from a degraded input sample, under some assumptions about the degradation model.
 In this review paper, we mainly focus on strategies for signal/image reconstruction, which is a classical task in the signal processing community.}
\label{fig:method_partioning}
\end{figure}

\begin{figure}
\centering
\includegraphics[width=0.5\textwidth]{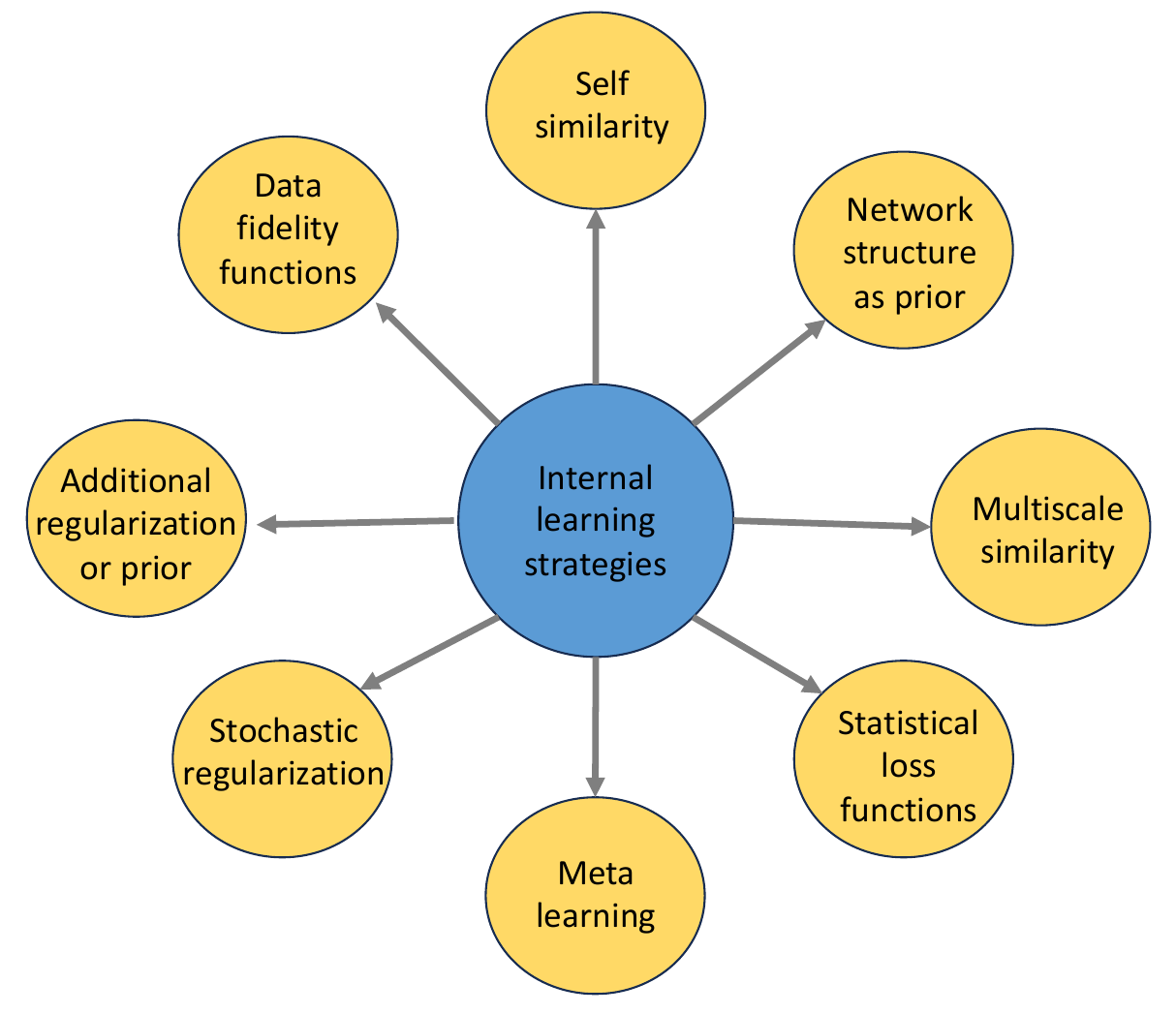}
 \caption{There are various strategies to perform internal learning. This figure highlights some of the key concepts / ingredients that are used in internal learning methods.}
\label{fig:internal_learning_concepts}
\end{figure}

In this review paper, we present a taxonomy for the different learning approaches that utilize internal learning.
The basis level of separation between techniques is whether they are fully based on internal learning, i.e., exploit only the input sample $x_0$ that is given at test time, or whether they incorporate internal learning with learning that is based on external data, e.g., fine-tuning pretrained models at test-time using $x_0$.
The latter can be separated into offline training methods that require ground truth clean images and ``unsupervised" methods that do not require clean data. As will be highlighted, techniques that train DNNs without the need for clean data oftentimes can be readily adapted for internal learning. 
Within each of these subgroups, we will distinguish between techniques according to their assumptions about the observation model and the input sample, as well as according to algorithmic aspects such as the DNN architecture, loss function and regularization.
We will discuss how each technique addresses the main limitations of internal learning, such as the long inference run-time and the sensitivity to early stopping.
The categorization according to the use of external data and the dependency on the observation model is visualized in Fig.~\ref{fig:method_partioning}, using several representative methods, which are among the methods that are surveyed in this paper. For each such method we will mention its key internal learning ingredients. A list of such ingredients is displayed in  Fig.~\ref{fig:internal_learning_concepts}.






In Section \ref{sec:internal_only} we discuss methods that train a DNN from scratch using only a single example $x_0$.
Most of these techniques are closely related to the DIP framework \cite{ulyanov2018deep}, and essentially try to mitigate its limitations while keeping using a massive U-net-like architecture. The modifications include loss functions different than \eqref{eq:DIP_loss} and various regularization techniques that can boost the results \cite{liu2019image,Zukerman20BPDIP}, be less sensitive to accurate early stopping \cite{Cheng2019Bayesian,Quan20Self2Self,wang2023early}, or both \cite{Metzler:2019gsure,Abu-Hussein_2022_WACV}.
In this part, special attention will be given to the classical 
Stein Unbiased Risk Estimator (SURE) \cite{Stein:1981vf} and its generalization \cite{eldar2008generalized} (GSURE), which provides a formula that estimates the MSE of $h(\cdot;\theta)$ w.r.t.~the latent $x_{gt}$ (independent of $x_{gt}$).
Unlike the traditional least squares loss whose minimization can eventually lead to fitting noise, the SURE criterion includes a term that regularizes the optimization and resolves this issue. 
Many recent works utilize (G)SURE \cite{soltanayev2018training,Zhussip_2019_CVPR,zhussip2019extending,kim2020unsupervised,Abu-Hussein_2022_WACV}. They demonstrate how concepts used in signal processing aid in self supervision.

In addition to methods that use overparameterized DNN architectures that are similar to DIP, we will present other internal learning techniques, such as zero-shot super-resolution (ZSSR) \cite{shocher2018zero} and deep decoder \cite{DeepDecoder} that eliminate the need for early stopping by using DNNs with less parameters. 
We also present approaches for learning generative models from a single image \cite{shaham2019singan,nikankin2022sinfusion,kulikov2023sinddm}.

It is worth mentioning that ZSSR, which was published concurrently with DIP, dubbed the term deep internal learning. Furthermore, as will be detailed below, the mechanism of these two methods differs beyond their architectures. Specifically, DIP  exploits the signal prior that is implicitly imposed by the DNN during unsupervised training (mapping random noise $z$ to the observations $x_0$). On the other hand, ZSSR explicitly exploits the across-scale similarity of signal/image patterns via self-supervised training (mapping a lower resolution version of $x_0$ to $x_0$).


\begin{table*}[!t]
\caption{Internal learning methods that do not use external data}
\label{table:internal_only}
\begin{center}
\begin{tabular}{ |p{1.7cm}|p{2.2cm}|p{2cm}|p{2.3cm}|p{5.7cm}| p{1.7cm}| } 
 \hline
 Method & Given info. & Structure & Input & Loss w.r.t. $\theta$ & Regularization \\ 
 \hline
 \hline
 DIP (denoise or JPEG) \cite{ulyanov2018deep} & noisy (or JPEG artifact) $x_0$ & encoder-decoder (conv) $h(\cdot;\theta)$ & uniform noise $z$ & $\| h(z;\theta) - x_0 \|_2^2$ & structure of $h(z;\theta)$ \\ 
  \hline
  DIP (super resol.) \cite{ulyanov2018deep} & low resolution $x_0$, $d(\cdot)$ downsamp. operator & encoder-decoder (conv) $h(\cdot;\theta)$ & uniform noise $z$ & $\| d(h(z;\theta)) - x_0 \|_2^2$ & structure of $h(z;\theta)$ \\ 
  \hline
  DIP (inpainting) \cite{ulyanov2018deep} & masked $x_0$, $m$ binary mask & encoder-decoder (conv) $h(\cdot;\theta)$ & uniform noise $z$ & $\| m \odot (h(z;\theta) - x_0) \|_2^2$ & structure of $h(z;\theta)$ \\ 
  \hline
 DIP (flash - no flash) \cite{ulyanov2018deep} & no flash $x_0$ \& flash $x_1$ & encoder-decoder (conv) $h(\cdot;\theta)$ & flash image $x_1$ & $\| h(x_1;\theta) - x_0 \|_2^2$ & structure of $h(z;\theta)$ \\ 
  \hline
  Double-DIP \cite{DoubleDIP} & image $x_0$ to separate & encoder-decoder (conv) $h(\cdot;\theta)$, $h(\cdot;\theta_1)$, $h(\cdot;\theta_2)$ & uniform noise $z$, $z_1$, $z_2$ (temporal consistency for video) & $\|h(z;\theta) \odot h(z_1;\theta_1) + (1-h(z;\theta)) \odot h(z_2;\theta_2) - x_0 \|_2^2 + L_{Ex}( h(z_1;\theta_1), h(z_2;\theta_2)) + L_{Reg}(h(z;\theta))$ & structure of $h(z;\cdot)$, min corr, task prior \\
  \hline
  DIP-SGLD \cite{Cheng2019Bayesian} & noisy or masked $x_0$ & encoder-decoder (conv) $h(\cdot;\theta)$ & Gaussian noise $z$ & $\| h(z, \theta) - x_0 \|_2^2$ with stochastic gradient Langevin dynamics w.r.t. $(z, \theta)$ & structure of $h(z;\theta)$, weight decay \\ 
  \hline
 Deep Decoder \cite{DeepDecoder} & noisy or low resol. or masked $x_0$, $f(\cdot)$ imaging model & decoder only (no conv) & random tensor $z$ & $\| f(h(z;\theta)) - x_0 \|_2$ & structure of $h(z;\theta)$, underparam. $\theta$ \\
 \hline
  DIP-TV \cite{liu2019image} & noisy or blurred $x_0$, $f(\cdot)$ imaging model & encoder-decoder (conv) $h(\cdot;\theta)$ & random noise $z$ & $\| f(h(z, \theta)) - x_0 \|_2^2 + L_{TV}(h(z;\theta))$ & structure of $h(z;\theta)$, total variation (TV) \\ 
 \hline
 DIP-SURE \cite{Jo_2021_ICCV} & noisy $x_0$ and its noise level $\sigma^2$ & encoder-decoder (conv) $h(\cdot;\theta)$ & $x_0+\gamma$, Gaussian noise $\gamma$ with uniform random var. & SURE $\| h(x_0+\gamma;\theta) - x_0 \|_2^2 + 2 \sigma^2 \mathrm{div} (h(x_0+\gamma;\theta))$, $\mathrm{div}$ divergence & structure of $h(z;\theta)$ \\ 
 \hline
  BP-TV \cite{Zukerman20BPDIP} & blurred $x_0$, $f(\cdot)$ imaging model & encoder-decoder (conv) $h(\cdot;\theta)$ & random noise $z$ & $\| (ff^T)^{-1/2} \{ f(h(z, \theta))-x_0 \} \|_2^2+L_{TV}(h(z;\theta))$ & structure of $h(z;\theta)$, total variation (TV) \\ 
  \hline
  PnP-DIP \cite{Sun21PnP} & masked / low resol. / noisy $x_0$, $f(\cdot)$ imaging model & encoder-decoder (conv) $h(\cdot;\theta)$ & random noise $z$ & $\| f(h(z, \theta)) - x_0 \|_2^2 + L_{Reg}(x)$ such that $x = h(z;\theta)$ & structure of $h(z;\theta)$, regularizers \\ 
  \hline
  DIP-RED \cite{mataev2019deepred} & noisy / low resol. / blurred $x_0$, $f(\cdot)$ imaging model & encoder-decoder (conv) $h(\cdot;\theta)$ & random noise $z$ & $\| f(h(z, \theta)) - x_0 \|_2^2 + \lambda x'(x-g(x))$ such that $x = h(z;\theta)$, $g$ denoiser & structure of $h(z;\theta)$, denoiser $g$ \\ 
  \hline
 DIP-SURE \cite{Metzler:2019gsure} & noisy $x_0$ and its noise level $\sigma^2$ & encoder-decoder (conv) $h(\cdot;\theta)$ & $x_0$ or $x_0+\gamma$, Gaussian noise $\gamma$ & SURE $\| h(x_0;\theta) - x_0 \|_2^2 + 2 \sigma^2 \mathrm{div} (h(x_0;\theta))$ or $\| h(x_0+\gamma;\theta) - x_0 \|_2^2$ & structure of $h(z;\theta)$ \\ 
 \hline
P-GSURE PnP \cite{Abu-Hussein_2022_WACV} & blurred / low resol. $x_0$, $f(\cdot)$ imaging model & encoder-decoder (conv) $h(\cdot;\theta)$ & $u=f^T(x_0)$ & P-GSURE $\| P_f h(u;\theta) \|_2^2 - 2h^T(u;\theta) f^\dagger(x_0) + 2 \mathrm{div}_u (P_f h(u;\theta)) + L_{Reg}(x)$ s.t. $x = h(u;\theta)$
$P_f$ projector for $f$, $f^\dagger$ pseudo-inverse of $f$ & structure of $h(z;\theta)$, regularizers \\ 
 \hline
 Self2Self \cite{Quan20Self2Self} & noisy $x_0$ & encoder-decoder (conv) $h(\cdot;\theta)$, dropout decoder & $\hat{x}_i = b_i \odot x_0$, Bernoulli masks $b_i$ & $\sum_i \| m_i \odot \{ h(\hat{x}_i;\theta) - m_i \odot x_0 \} \|_2^2$, $m_i=1-b_i$ & structure of $h(u;\theta)$ \\ 
  \hline
   ZeroShot-Noise2Noise \cite{mansour2023zero} & noisy $x_0$ & small fully CNN $h(\cdot;\theta)$ & $x_0$, $d_1(x_0)$, $d_2(x_0)$ where $d_i(\cdot)$ downsamp. operators & $\sum_{i=1}^2 \| d_i(x_0) - h(d_i(x_0);\theta) - d_{j(i)}(x_0) \|_2^2 + \| d_i(x_0) - h(d_i(x_0);\theta) - d_{i}(x_0 - h(x_0;\theta)) \|_2^2 $ where $j(1)=2, j(2)=1$ & underparam. CNN $\theta$, symmetric losses \\ 
  \hline
  ZSSR \cite{shocher2018zero} & low resolution $x_0$, $d(\cdot)$ downsamp. operator & small fully CNN $h(\cdot;\theta)$ & patches of low resol. image $P_i x_0$ & $\sum_i \| h(d(P_i x_0);\theta) - P_i x_0 \|_2^2$ & underparam. CNN $\theta$ \\ 
  \hline
  SinGAN \cite{shaham2019singan} & sample image $x_0$, $x_i = d_i(x_0)$ down sampling, $i$th level & a pyramid of CNNs $h(\cdot;\theta_i)$ & multi-scale random noises $z_0,\ldots,z_N$ &
  $L_{adv}( h(z_i, u(\hat{x}_{i+1});\theta_i), x_i) + \| h(0,u(\hat{x}_{i+1});\theta_i) - x_i \|_2^2$, $u$ upsampling, $L_{adv}$ adversarial loss with discriminator & pyramidal CNNs \\ 
  \hline
\end{tabular}
\end{center}
\end{table*}

\begin{table*}[!t]
\caption{Methods that incorporate internal learning with models trained via external data}
\label{table:internal_and_external}
\begin{center}
\begin{tabular}{ |p{1.7cm}|p{2.2cm}|p{2cm}|p{2.3cm}|p{5.7cm}| p{1.7cm}| } 
\hline
Method & Given info. & Structure & Input & Loss w.r.t. $\theta$ & Regularization \\ 
\hline
\hline
Noise2Noise \cite{noise2noise} & independent noisy $x_0$, $x_1$ (same GT) & any $h(\cdot;\theta)$ & noisy $x_0$ & $\| h(x_0;\theta) - x_1 \|_2^2$ & - \\ 
\hline
Noise2Void \cite{krull2019noise2void} & noisy image $x_0$ & any $h(\cdot;\theta)$ & $\hat{x}_i = m_i \odot x_0$, $m_i = 1 - b_i$ & $\sum_i \| b_i \odot \{ h(\hat{x}_i;\theta) - b_i \odot x_0 \} \|_2^2$, $b_i$ single pixel mask & - \\
\hline  
Noise2Self \cite{batson19noise2self} & noisy image $x_0$ & any $h(\cdot;\theta)$ & $\hat{x}_i = m_i \odot x_0$, $m_i = 1 - b_i$ & $\sum_i \| b_i \odot \{ h(\hat{x}_i;\theta) - b_i \odot x_0 \} \|_2^2$, $b_i$ binary partition mask & a single large image \\
\hline
 SURE \cite{soltanayev2018training} & Gaussian noisy $x_0$ and noise level $\sigma^2$ & any $h(\cdot;\theta)$ & noisy $x_0$ & $\| h(x_0;\theta) - x_0 \|_2^2 + 2 \sigma^2 \mathrm{div} (h(x_0;\theta))$, $\mathrm{div}$ divergence & fine-tuning \\
 \hline
 CS-SURE \cite{Zhussip_2019_CVPR} & CS measurement $x_0$ ($\in \mathcal{C}^M$), CS model $f(\cdot)$ & any $h(\cdot;\theta)$ & $\hat{x}+f^T(\hat{z})$, $\hat{x}$ image estimate & $\| h(\hat{x}+f^T(\hat{z});\theta) - \hat{x}+f^T(\hat{z}) \|_2^2 + 2 \hat{\sigma}^2 \mathrm{div} (h(\hat{x}+f^T(\hat{z});\theta))$, 
 $\hat{z} = x_0 - f(\hat{x}) + \hat{z} \mathrm{div} (\hat{x}+f^T(\hat{z}))/M$,
 $\hat{\sigma} = \| \hat{z} \|_2/\sqrt{M}$
 & fine-tuning \\ 
 \hline
 eSURE \cite{zhussip2019extending} & Gaussian noisy $x_0$ and noise level $\sigma^2$ & any $h(\cdot;\theta)$ & noisy $x_0 + z$, $z$ Gaussian noise & $\| h(x_0 + z;\theta) - x_0 \|_2^2 + 2 \sigma^2 \mathrm{div} (h(x_0+z;\theta))$, $\mathrm{div}$ divergence & fine-tuning \\
 \hline
 CT-PURE \cite{kim2020unsupervised} & CT measurement $x_0$ & any $h(\cdot;\theta)$ & Poisson noisy CT measurement $x_0$ & $\| h(x_0;\theta) - x_0 \|_2^2 - \langle 1, x_0 \rangle + 2 \langle x_0, \partial h(x_0;\theta) \rangle$, inner product $\langle \cdot, \cdot \rangle$
 & fine-tuning \\ 
 \hline
MZSR \cite{soh2020meta} & low resolution $x_0$, $d(\cdot)$ downsamp. operator & any $h(\cdot;\theta)$ & low resolution $x_0$ & Meta-test $\| h(d(x_0);\theta) - x_0 \|_2^2$ & meta learning in kernels \& meta test \\
\hline
MLSR \cite{park2020fast} & low resolution $x_0$, $d(\cdot)$ downsamp. operator & any $h(\cdot;\theta)$ & low resolution $x_0$ & Meta-test $\| h(d(x_0);\theta) - x_0 \|_2^2$ & meta learning in images \& meta test \\
\hline
IAGAN \cite{IAGAN} & 
degraded $x_0$, $f(\cdot)$ forward model
& pre-trained GAN $h(\cdot;\theta)$ & 
degraded $x_0$ 
& $\| f( h(z;\theta) ) - x_0 \|_2^2$ w.r.t. $\theta$ and $z$. Then, $\hat{x} = h(z^*;\theta^*)$; Optional in noiseless settings: $\hat{x}_{bp} = \hat{x} + f^\dagger(x_0 - f(\hat{x}))$ & fine-tuning \& optional back-proj. \\
\hline
IDBP-CNN \cite{tirer2019super} & noisy, low resol. $x_0$, $f(\cdot)$ forward model & pre-trained denoiser $h(\cdot;\theta)$ & noisy low resolution $x_0$ & $\| h(x_0 + z;\theta) ) - x_0 \|_2^2$ where $z$ is a random vector, then for $\hat{x} = h(\hat{x};\theta^*)$, $\hat{x}_{bp} = \hat{x} + f^\dagger(x_0 - f(\hat{x}))$ & fine-tuning \& back-projection \\
\hline
DGP \cite{pan2020dgp} & degraded $x_0$, $f(\cdot)$ forward model & pre-trained GAN $h(\cdot;\theta)$ & degraded $x_0$ & $\| f( h(z;\theta) ) - x_0 \|_2^2$ w.r.t. progressively partial $\theta$ and full $z$. Then, $\hat{x} = h(z^*;\theta^*)$ & progressive fine-tuning \\
\hline
GainTuning \cite{mohan2021adaptive} & noisy $x_0$ & pre-trained denoiser $h(\cdot;\theta)$ & noisy $x_0$ & the loss of Noise2Void or SURE, 
but w.r.t. partial $\theta$ (normalization layers) & partial fine-tuning \\
\hline
ADIR \cite{abu2022adir} & degraded $x_0$, $f(\cdot)$ forward model & pre-trained diffusion's denoiser $h(\cdot;\theta)$ 
& degraded $x_0$ & $\| h(\tilde{x}^k + z;\theta) ) - \tilde{x}^{(k)} \|_2^2$ where $z$ is a random vector
and $\{\tilde{x}^k\}$ are $x_0$ and/or k-NNs of $x_0$ in neural embedding space from an external dataset & fine-tuning \\
\hline
\end{tabular}
\end{center}
\end{table*}

The main limitation of ``pure" internal learning -- which uses only the single observed image for training DNNs -- is the potential performance drop due to not exploiting the massive amount of external data that is available for many tasks. 
This led to the idea of incorporating offline external and test-time internal learning to get the best of both worlds \cite{tirer2019super}.

In Section~\ref{sec:test_time_adaptation}, we discuss different methods for test-time fine-tuning of DNNs that have been already pretrained offline.
Focusing on image restoration, we present methods that adapt deep priors, such as CNN denoisers \cite{tirer2019super}  and generative adversarial networks (GANs) \cite{goodfellow2014generative} as initiated in \cite{IAGAN} and its follow up works \cite{pan2020dgp,Daras2021ILO}. All of these methods may be plugged into quite general frameworks that can tackle different restoration tasks. In other words, they allow {\em flexibility} in the observation/forward model $f(\cdot)$, contrary to offline trained task specific DNNs.



When adapting pretrained models to the test image at hand, special care should be given to make sure that useful semantics / patterns that have been captured offline will not be overridden or `forgotten' during the test time optimization.
This risk which is typically addressed by early stopping, can be further mitigated by optimizing only a small set of the pretrained model's parameters  \cite{mohan2021adaptive}.
Another limitation of test-time tuning that is important to address is the additional inference run-time that is added to the pretrained model.
To mitigate this issue, several 
meta-learning approaches have been used \cite{soh2020meta,park2020fast}.

\section{Learning using a single input example}
\label{sec:internal_only}

In this section, we discuss different methods that train a DNN from scratch using only a single example $x_0$.
Our reference point will be the DIP approach \cite{ulyanov2018deep}, which has been described in Section~\ref{sec:overview}.
In Section~\ref{sec:internal_only_arch} we focus on architectural variations of DIP, and in Section~\ref{sec:internal_only_loss} we focus on algorithmic variations, mainly in terms of the optimization objective (i.e., alternatives to the loss function in \eqref{eq:DIP_loss}).
These variations aim to improve the reconstruction accuracy of DIP or to mitigate its large inference time and sensitivity to early stopping.
The methods that are discussed in this section are summarized in Table~\ref{table:internal_only}.

\subsection{Architecture-based approaches}
\label{sec:internal_only_arch}

The core idea of DIP is that the network structure is an implicit signal prior.
The most widely used network structure for DIP $h(z; \theta)$ (cf. \eqref{eq:DIP_loss}) is the ``hourglass'' (also known as ``encoder-decoder'') architecture~\cite{ulyanov2018deep}, which is $h(z; \theta) = h_d( h_e(z; \theta_e); \theta_d )$ where $h_e(z; \theta_e)$ is the encoder whose outputs are the latent vector as well as skip connections, $h_d(\cdot; \theta_d)$ is the decoder whose output is the enhanced image, and $\theta$ is the network parameter vector, which is a concatenation of the encoder network parameter vector $\theta_e$ and the decoder network parameter vector $\theta_d$. The encoder network consists of sets of convolution, batch normalization and non-linear activation with downsampling and the decoder network consists of sets of the same components as the decoder network, but replacing downsampling by upsampling. The encoder and decoder networks are additionally connected via skip connections at the same spatial resolution of features. 
Note that this architecture is highly overparameterized. In general, the widespread belief is that overparameterization facilitates optimization in deep learning. However, in our case, where only a single image $x_0$ is given, overparameterization also allows overfitting the noise and artifacts in $x_0$ after a large number of optimization iterations (cf. Fig~\ref{fig:dip_signal_before_noise}), which is undesired and is the core reason for accurate early stopping.

Different regularization structures have also been proposed as alternatives  to DIP's architecture. Deep Decoder (DD)~\cite{DeepDecoder} removed the encoder network and enforced a simple tensor product structure in the decoder network architecture $h_d(z; \theta_d)$. 
The limited capacity of DD compared to DIP, makes it robust to the stopping point of the optimization, at the price of performance drop compared to DIP with optimal early stopping (which may not be feasible in practice).
A decoder only regularization structure for dynamic MRI was also investigated in \cite{yoo2021time} where a low-dimensional manifold structure in $z$ encodes the temporal variations of images unlike a static, random vector $z$ in most DIP works. 

\begin{figure}
    \centering
  \includegraphics[width=200pt]{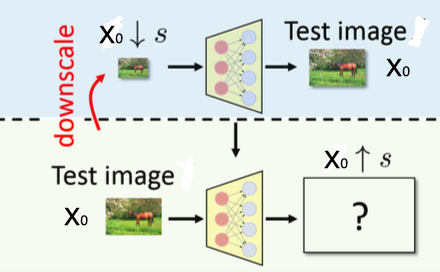} 
  \caption{``Zero-shot super-resolution" (ZSSR) approach: given a known downsampling model $f(\cdot)=(\cdot)\downarrow_s$ and the low-resolution observation $x_0={(x_{gt})\downarrow_s}$, internal learning is performed by training a moderate size CNN to map ${(x_0)\downarrow_s}$ to $x_0$ (with patch extraction and augmentations). After the optimization phase, the network is applied on $x_0$ to estimate $x_{gt}$. 
  Figure taken from \cite{shocher2018zero} \tomt{and used by permission of the authors}.}
\label{fig:zssr_sketch}
\end{figure}

Similarly to DD, another concise network structure has been proposed for zero shot super resolution (ZSSR) \cite{shocher2018zero}.
This approach focuses on the super-resolution task, namely, $f: \mathbb{R}^n \to \mathbb{R}^m$ in the observation model \eqref{eq:observation_model} is a downsampling operation $d(\cdot)$ (with $m \ll n$), which is a composition of filtering with an arbitrary low-pass kernel and subsampling. 
The DNN $h(\cdot,\theta)$ used in \cite{shocher2018zero} is a relatively simple 8-layer fully convolutional network. Since the dimension of the output of this DNN is the same as the dimension of the input to the first convolutional layer, before reaching the DNN, the input image goes through bicubic upsampling (regardless of the low-pass kernel in $d(\cdot)$, which can be arbitrary), such that the network's output is of higher dimension and can estimate the unknown high resolution image.  
Training this DNN is different from DIP and DD. The low-resolution $x_0$ is downsampled itself to create an even lower resolution image, $d(x_0)$, and then a network is trained to reconstruct from it the given input image, $x_0$, which is of higher resolution. Specifically, the loss function used in ZSSR is given by
\begin{align}
\label{eq:zssr}
    \sum_i \| h(P_i d(x_0);\theta) - P_i x_0 \|_2^2,
\end{align}
where $P_i$ denotes patch extraction and the sum goes over the different patches (including those obtained by various augmentations).
After the optimization phase, the trained DNN is applied to the original low-resolution image $x_0$ to produce its higher-resolution version, which is an estimate of $x_{gt}$.
The ZSSR scheme is presented in Fig.~\ref{fig:zssr_sketch}. 
A similar technique has also been used for learning to improve 3D shapes \cite{Metzer21SelfSample}. 
The idea behind this technique is that signals like natural images have recurring patterns even across scales of resolution and not only within the same scale. 

Further extension of the DIP was proposed for decomposing images into their basic components by exploiting the representation power of the DIP on low-level statistics of an image \cite{DoubleDIP} (dubbed Double-DIP). The network structure of Double-DIP consists of two DIP networks or $m \odot h(z_1; \theta_1) + (1-m) \odot h(z_2; \theta_2)$ such that $m = h(z_m; \theta_m)$. Then, two basic components in a complex image can be represented with $h(z_1; \theta_1)$ and $h(z_2; \theta_2)$, respectively, and the separation mask $m$ will be estimated with $h(z_m; \theta_m)$ under some assumptions on $m$ such as binary mask (for segmentation problems) or smooth mask (for dehazing tasks). In this case, the data fidelity term in the loss that is used is given by
\begin{align}
\|h(z;\theta) \odot h(z_1;\theta_1) + (1-h(z;\theta)) \odot h(z_2;\theta_2) - x_0 \|_2^2, 
\end{align}
to which two additional terms are added: $L_{Ex}( h(z_1;\theta_1), h(z_2;\theta_2))$ that reduces the correlation between the gradients of the two components, and a task-specific regularization $L_{Reg}(h(z;\theta))$. 

Finally, while our paper mostly focuses on using internal learning for classical signal and image processing tasks, which aim to estimate $x_{gt}$ given $x_0$, we note that internal learning has also been used for generative modeling: synthesizing new samples given $x_{gt}$.

A prominent work in this line is SinGAN \cite{shaham2019singan}, which is based on advances in generative adversarial networks (GANs) \cite{goodfellow2014generative}.
In GANs, the goal is to train a generator network, $h(\cdot;\theta):\mathbb{R}^k \to \mathbb{R}^n$, to map a low-dimensional Gaussian vector $z \in \mathbb{R}^k$ to an image in $\mathbb{R}^n$ such that another trainable network, the discriminator (or critic) $c(\cdot;\tilde{\theta}):\mathbb{R}^n \to [0,1]$ (with 0=fake and 1=real), fails to distinguish between the generator's outputs and images that belong to the real training data. Commonly, the networks are trained by alternating minimization w.r.t.~$\theta$ and maximization w.r.t.~$\tilde{\theta}$ of the adversarial loss:
$$
L_{adv}( h(z;\theta), x_{gt}) = \log c(x_{gt}) + \log (1-c(h(z;\theta);\tilde{\theta})),
$$
where $z$ and $x_{gt}$ are drawn at each optimization iteration from $\mathcal{N}(0,I_k)$ and from the training data, respectively. 

Despite the possibility of using a generator network similar to the Unet-like architecture of DIP, 
\tomt{the generator network used in SinGAN is different.}  
Instead of having only a single input $z$ to the network with dimensions similar to $x_{gt}$, a multi-resolution approach is  used. An image pyramid of $x_{gt}$ using a down sampling operation $x_{gt,i} = d_i(x_{gt})$ ($i$ denotes the resolution level), as well as a pyramid of CNNs $h_i(\cdot,\cdot;\theta_i)$, where the first argument is the latent (noise) vector and the second argument is the output of the previous lower resolution level $i+1$.
Starting from the lowest resolution and gradually reaching to the original resolution, $h_i(\cdot,\cdot;\theta_i)$ is trained to map random input $z_i$ of the same dimension as $x_{gt,i}$, conditioned on an upsampled version of the output of the lower level $u(\hat{x}_{i+1})$, by minimizing a loss function of the form
$$
L_{adv}( h_i(z_i, u(\hat{x}_{i+1});\theta_i), x_{gt,i}) + \| h_i(0,u(\hat{x}_{i+1});\theta_i) - x_{gt,i} \|_2^2,
$$
where $L_{adv}$ is an adversarial loss with patch-based discriminator and the second term is a reconstruction loss term.
In \cite{shaham2019singan}, it has been shown that the internally learned generator can produce perceptually pleasing variations of the given $x_{gt}$.

Recently, score/diffusion-based generative models \cite{song2019generative,ho2020denoising} have been shown to be a powerful alternative to GANs.
In this approach, during the training a U-net Gaussian denoising network, conditioned on the noise level (or equivalently ``time index") is trained for a large range of noise levels.
In contrast to the Unet-like network used in DIP, the input to the network used for score/diffusion-based generation is the noisy images $x_{gt}+e$ (where $e$ is controlled as this is supervised training) and at the network's low-resolution levels there is usage of self-attention that allows capturing global semantics.
In inference time, new images are generated by initializing a noise image $x_T$ and, iteratively, given $x_t$ generating the next image $x_{t-1}$ by denoising and adding synthetic noise, both with decreasing noise levels, until reaching $x_0$ that resembles a sample from the data distribution.
Utilizing a multi-scale approach, similar to SinGAN, it has been recently shown that this score/diffusion-based sampling approach can be used with a denoiser trained on a single ground truth image $x_{gt}$ for generating its variations \cite{kulikov2023sinddm,nikankin2022sinfusion}.


\subsection{Optimization-based approaches}
\label{sec:internal_only_loss}

Most follow-up works of DIP do not modify its DNN architecture much, but rather try to improve its performance or mitigate its limitations by modifying network optimization.

A natural way to mitigate DIP tendency to fit the observation noise, and potentially to improve the reconstruction performance,
is by utilizing regularization techniques.
The authors of \cite{Cheng2019Bayesian} argued that merely adding $\ell_2$-norm regularization for the network's parameters $\theta$ (i.e., weight decay) is insufficient for preventing fitting the observations' noise. Instead, they proposed a DIP-SGLD approach, based on a technique known as stochastic gradient Langevin dynamics (SGLD), which can be motivated from a Bayesian point of view. In the method, a different noise realization is added to $\Delta \theta$, the gradient update of $\theta$, in each optimization iteration $t$:
\begin{align}
    \Delta \theta_t + \eta_t,
\end{align}
where $\eta_t \sim \mathcal{N}(0,\epsilon I)$ with an hyperparameter $\epsilon$ that obeys $\sum_t \epsilon_t = \infty$ and $\sum_t \epsilon_t^2 < \infty$. 
Empirically, it is then demonstrated that the need for accurate early stopping is spared.

Focusing on the denoising task, i.e., $f=I$ in the observation model \eqref{eq:observation_model}, 
another technique that regularizes the optimization via stochasticity is Self2Self \cite{Quan20Self2Self}. In this method, multiple random Bernoulli masks $\{b_i\}$ are zero pixels of the input image $\hat{x}_i = b_i \odot x_0$, and a loss function of the form
\begin{align}
\sum_i \| (1-b_i) \odot \{ h(\hat{x}_i;\theta) - (1-b_i) \odot x_0 \} \|_2^2
\end{align}
is used for optimizing the network's parameters $\theta$ to fill the missing pixels in all masked scenarios jointly.
Moreover, dropout regularization is used in the optimization of the decoder part of $h(\cdot;\theta)$.
This approach is not sensitive to an early stopping in the denoising task, and boosts the result (compared to DIP) if the ensemble of estimates, associated with the different masks, is aggregated.

The loss function of Self2Self can be understood as a generalization of the loss used in a related denoising method named Noise2Void \cite{krull2019noise2void}:
\begin{align}
\label{eq:noise2void}
    \sum_i \| b_i \odot \{ h(\hat{x}_i;\theta) - b_i \odot x_0 \} \|_2^2,
\end{align}
where $b_i$ deterministically chosen as a single pixel mask (i.e., erases the $i$th pixel) and the sum goes over the image patches.
Originally, this method was not proposed for internal learning, but rather for training a denoiser based on a dataset of noisy images without associated clean ground truth versions (and without multiple realizations of noise per image, contrary to its predecessor Noise2Noise \cite{noise2noise}).
Thus, the objective in \eqref{eq:noise2void} is further summed over the training samples.
However, note that since this loss does not require knowledge of $x_{gt}$, \tomt{it can be used} for internal learning as well, as later demonstrated in a follow-up work, Noise2Self \cite{batson19noise2self}, for a large single image.
All these methods \cite{Quan20Self2Self,krull2019noise2void,batson19noise2self} exploit the prior knowledge that there is a dependency between the intensity levels of neighboring pixels in clean natural images. This is in sharp contrast with the characteristics of noise, under the assumption that the noise distribution is independent per pixel.
Therefore, fitting the noise is mitigated by masking pixels.

Other regularization techniques include adding a regularization term to the loss function stated in \eqref{eq:DIP_loss}.
In the signal processing community, the total variation (TV) criterion is a prominent regularizer that is based on the observations that many signals are piece-wise constant, and thus their gradients are sparse (have many zeros). Specifically, for a two-dimensional signal $x$, the anisotropic TV regularizer is given by
$$
L_{TV}(x) = \sum_{i,j} |x_{i+1,j}-x_{i,j}| + |x_{i,j+1}-x{i,j}|. 
$$
In \cite{liu2019image} it has been shown that DIP-TV, a variant of DIP where TV regularization is added to the loss function:
\begin{align}
    \| f(h(z, \theta)) - x_0 \|_2^2 + \lambda L_{TV}(h(z;\theta))
\end{align}
yields performance gains. Yet, since natural images are not really piece-wise constant, this boost is obtained with a small regularization  parameter $\lambda$, and thus early stopping is still required, as shown in \cite{Zukerman20BPDIP}.

Maintaining the TV term, the authors of \cite{Zukerman20BPDIP} suggested another modification to the loss function. Specifically, motivated by \cite{tirer2018image,tirer2020back}, they replaced the least squares data-fidelity term \eqref{eq:DIP_loss} with a specific type of weighted least squares, dubbed the back-projection (BP) term, given by
\begin{align}
\label{eq:bp_term}
\| (ff^T+\epsilon I)^{-1/2} \{ f(h(z, \theta))-x_0 \} \|^2,
\end{align}
where the forward operator $f$ is assumed to be linear and some diagonal regularization is used when $ff^T$ is not well conditioned.
In \cite{tirer2018image,tirer2020back} it has been shown that using BP rather than least squares yields better results in the low-noise regime and accelerates optimization. Recently, the concept has been generalized to the high-noise regime by smoothly shifting from BP to least squares along the optimization \cite{garber2023image}.
In agreement with previous observations,
the proposed BP-TV in \cite{Zukerman20BPDIP} has been shown to yield better results 
than DIP-TV 
in the low-noise regime and, importantly, with much fewer optimization iterations. This is especially advantageous for internal learning methods, as the optimization is done in inference (test) time. Yet, the need for accurate early stopping remains.

Instead of utilizing an explicit regularization term, a promising direction in recent years is to use off-the-shelf/pretrained denoisers to impose the signal's prior, which mostly follows the PnP \cite{venkatakrishnan2013plug} and RED \cite{RED} approaches.
Let $g : \mathbb{R}^n \to \mathbb{R}^n$ denote an off-the-shelf denoiser.
PnP conceptually adds a prior term $s: \mathbb{R}^n \to \mathbb{R}$ to the loss function, but then replaces its prox operator in proximal optimization algorithms (ADMM, prox-grad method, etc.) by the denoiser $g$.
RED, on the other hand, adds to the gradient of the data-fidelity term a (scaled) gradient of the implicit prior $s$, which takes the form of $x \mapsto x - g(x)$.
Applying these approaches to impose regularization via modern denoisers has demonstrated better results than using classical techniques like TV.
As a constructive example, a scheme of prox-grad-based PnP is given in Section~\ref{sec:test_time_adaptation_incorp}, where internal learning is used for fine-tuning a pretrained denoiser.

Advances in PnP and RED have been utilized also in internal learning.
The authors of \cite{mataev2019deepred} proposed DIP-RED, as a method that boosts the result of the plain DIP via existing denoisers.
Similarly, \cite{Abu-Hussein_2022_WACV} and \cite{Sun21PnP} improved the results of DIP via a PnP approach.
Yet, PnP/RED adaptations of DIP have some disadvantages such as increasing the inference time due to alternating between multiple network optimization (plain DIP) and denoiser applications, as well as not addressing the stopping time issue.

\begin{tcolorbox}[breakable,colback=blue!5!white,colframe=blue!75!black,title=Box 2: Internal Learning Using SURE Criterion and DNN parameterization]

SURE based approaches \cite{Metzler:2019gsure,soltanayev2018training,eldar2008generalized} tackle the denoising task: estimating $x_{gt}$ from $x_0 = x_{gt}+e$, where $e \sim \mathcal{N}(0,\sigma^2 I)0$.
Similarly to DIP, they utilize the implicit prior induced by U-net parameterization of the estimate $\hat{x}=h(u,\theta)$, but unlike DIP, the input to the DNN is a sufficient statistic of the problem $u=x_0$ rather than noise, and instead of minimizing the plain least squares loss $\|h(x_0,\theta)-x_0\|^2$, they minimize the estimate of the MSE given by:
\begin{align*}
    &\min_\theta SURE(\hat{x}(u,\theta)) = \\ &\min_\theta \,\, \|h(x_0;\theta) - x_0 \|^2 + 2\sigma^2 \mathrm{div} (h(x_0;\theta)).
\end{align*}
Thus, the key difference between plain DIP and SURE is penalizing the optimization according to the network divergence (essentially, the trace of its Jacobian). 
This regularization hardens fitting the noise as it prevents sensitivity to pixel-wise changes in $x_0$.

The figures below, taken from \cite{Metzler:2019gsure} \tomt{and used by permission of the authors}, show the advantages of minimizing the SURE criterion instead of a typical least squares criterion in image denoising.
On the left, the network's divergence is not controlled (as in DIP), and the (normalized) MSE increases since the DNN starts fitting the noise.
On the right, the SURE criterion controls the divergence and an increase in the MSE is prevented --- and thus, no accurate early stopping is required.

\vspace{3mm}

\includegraphics[width=0.95\linewidth]{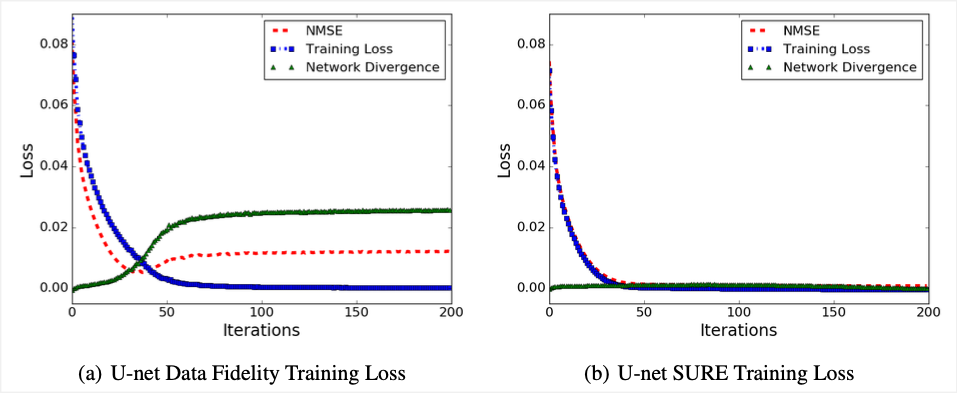}

See the text for details on the generalization of SURE beyond the image denoising setting.

\end{tcolorbox}

We next introduce a family of methods that overcome the requirement for early stopping, while still being based on an explicit analytical objective.
To introduce the core idea behind these methods, let us focus on the denoising task, i.e., $f=I$ in the observation model \eqref{eq:observation_model} and $e$ is white Guassian noise with known variance $\sigma^2$ (though the following discussion can be generalized to other exponential noise distributions). Let $\hat{x}(x_0)$ denote an estimator of the clean $x_{gt}$. 
The plain least squares loss used in DIP \eqref{eq:DIP_poc} (equivalently \eqref{eq:DIP_loss} with $f=I$) can easily fit the noisy $x_0$, which is aligned with $\hat{x}(x_0)=x_0$ being the minimizer of $\|\hat{x}(x_0) - x_0\|^2$.
If, on the other hand, an oracle would have given us the MSE criterion 
$MSE(\hat{x}) = \mathbb{E} \|\hat{x}(x_0) - x_{gt}\|^2
$, 
where the expectation is taken over the noise in $x_0$, then no noise overfitting can occur (and obviously, we can expect better performance).
However, $x_{gt}$ is the unknown image that we need to estimate in the first place.

\tomt{In a foundational work \cite{Stein:1981vf},
Stein} proposed an unbiased risk (MSE) estimate, nowadays known as SURE:
\begin{align}
\label{eq:sure}
    SURE(\hat{x}) = -n\sigma^2 + \|\hat{x}(x_0) - x_0 \|^2 + 2\sigma^2 \mathrm{div} (\hat{x}(x_0)),
\end{align}
where the divergence operator reads as $\mathrm{div}(h(u)) = \sum_i \frac{\partial}{\partial u_i}[x(u)]_i$.
The unbiasedness of SURE reads as $\mathbb{E}[SURE(\hat{x})] = MSE(\hat{x})$.
Crucially in our case, the divergence term penalizes the estimator for being sensitive to $x_0$, which essentially hardens fitting the noise. 
To facilitate the usage of SURE, it is common to approximate the divergence term by
$$
\mathrm{div} (\hat{x}(x_0)) \approx \eta^T (\hat{x}(x_0+\epsilon\eta) - \hat{x}(x_0))/\epsilon
$$
with small $\epsilon>0$ and $\eta \sim \mathcal{N}(0, I)$.

In the past, the SURE criterion has been mostly used to tune only one or two parameters of an estimator. However, following the advances in deep learning, works have suggested utilizing SURE even for $\hat{x}(x_0)$ (over)parameterized by deep neural networks \cite{soltanayev2018training,Metzler:2019gsure}. In the context of internal learning, 
\cite{Metzler:2019gsure} proposed a DIP-SURE approach:
mitigating the problem of DIP fitting the noise by parameterizing $\hat{x}(x_0)$ by DIP's architecture $\hat{x}(x_0) = h(x_0;\theta)$, with the difference that the input is the observations $x_0$ rather than a drawn noise image $z$, and optimizing the DNN's parameters $\theta$ by minimizing $SURE(h(x_0;\theta))$ rather than the typical least squares term. 
In \cite{Jo_2021_ICCV}, the method was further improved by adding random perturbations to $x_0$ in the input and in the divergence term. 

In Box 2 we present figures that demonstrate the importance of the additional divergence term in SURE. Specifically, the increase in the divergence of the network is an indicator of fitting the noise in the observations. Thus, penalizing it, as done in SURE, resolves the need for accurate early stopping.

\begin{figure}[t]
    \centering
    \begin{subfigure}{0.45\linewidth}
        \centering
        \includegraphics[width=130pt, height = 100pt]{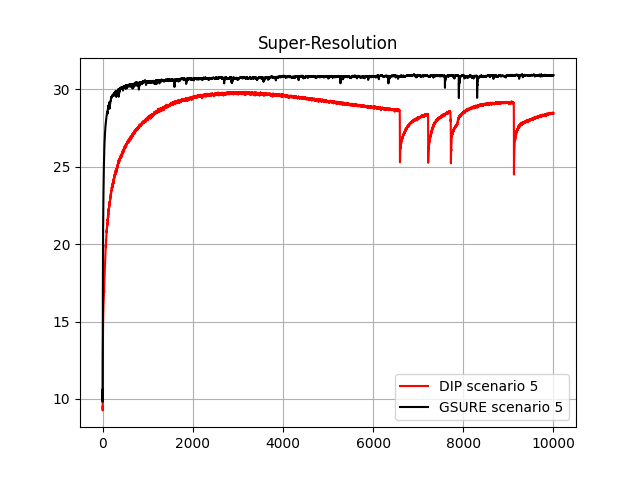}
        \caption{PSNR (Set5) vs.~iter.}
    \end{subfigure}
    \begin{subfigure}{0.45\linewidth}
        \centering
        \includegraphics[width=100pt, height = 100pt]{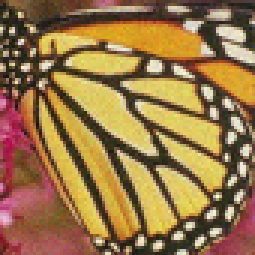}
    \caption{Noisy low-res. $x_0$}
        \vspace{0.5mm}
    \end{subfigure}
    \begin{subfigure}{0.45\linewidth}
    \vspace{2mm}
        \centering
        \includegraphics[width=100pt, height = 100pt]{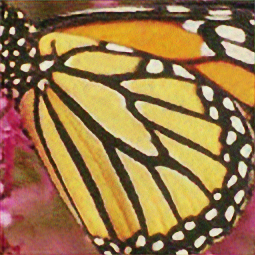}
        \caption{DIP}  
    \end{subfigure}
    \begin{subfigure}{0.45\linewidth}
    \vspace{2mm}
        \centering
        \includegraphics[width=100pt, height = 100pt]{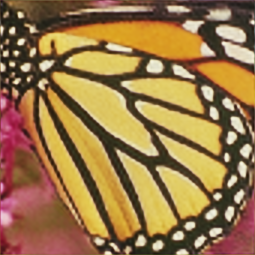}
        \caption{GSURE}       
    \end{subfigure}
    \begin{subfigure}{0.45\linewidth}
    \vspace{2mm}
        \centering
        \includegraphics[width=100pt, height = 100pt]{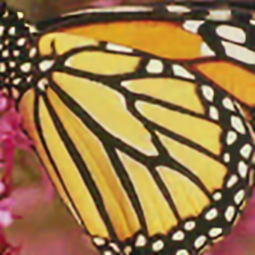}
        \caption{DIP-PnP}  
    \end{subfigure}
    \begin{subfigure}{0.45\linewidth}
    \vspace{2mm}
        \centering
        \includegraphics[width=100pt, height = 100pt]{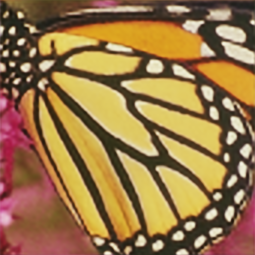}
        \caption{GSURE-PnP}       
    \end{subfigure}    
    \caption{Super-resolution x3 with a bicubic kernel and noise level of $\sqrt{10}/255$. (a) PSNR average over Set5 versus ADAM iteration number; (b) An observed image; (c) DIP recovery; (d) GSURE recovery; (e) DIP-PnP(BM3D) recovery; (f) GSURE-PnP(BM3D) recovery. As observed in (a), the DIP recoveries start fitting the noise at some iteration while the GSURE recoveries do not suffer from this issue. In this example, both DIP and GSURE benefit from an additional prior imposed by the plug-and-play BM3D denoiser.
    Figures are taken from \cite{Abu-Hussein_2022_WACV} \tomt{and used by permission of the authors}.}
    \label{fig:comparison_sr_scenario5}
\end{figure}

We now turn to discuss a more general observation model, specifically, the case of the linear forward operator $f$.
A generalized version of SURE (GSURE) suitable for this case has been derived in \cite{eldar2008generalized}:
\begin{align}
\label{eq:gsure}
    GSURE(\hat{x}) = &-\sigma^2\mathrm{Tr}(f^\dagger f^{\dagger T}) + \|f^\dagger f \hat{x}(u) - f^\dagger x_0 \|^2  \nonumber \\ 
    &+ 2\sigma^2 \mathrm{div} (f^\dagger f \hat{x}(u)),
\end{align}
where $f^\dagger$ denotes the pseudoinverse of $f$ and $u \in \mathbb{R}^n$ is a sufficient statistic (e.g., $u=f^T x_0$).
This expression is an unbiased estimate of the ``projected MSE", namely, the component of the signal in the row range of $f$: $\mathbb{E} \|f^\dagger f ( \hat{x}(u) - x_{gt})\|^2$.
Accordingly, \cite{Abu-Hussein_2022_WACV} proposed DIP-GSURE as an extension of DIP-SURE that can address tasks other than denoising (e.g., deblurring and super-resolution) and showed robustness to the stopping iteration.
The empirical faster convergence of minimizing \eqref{eq:gsure} than plain least squares objective is explained in \cite{Abu-Hussein_2022_WACV} by showing that minimizing GSURE is equivalent to minimizing the sum of the BP-term \eqref{eq:bp_term} with the divergence  term. The latter term is also the reason that no additional regularization is required when handling measurements at high noise levels. 
We note that \cite{Abu-Hussein_2022_WACV} also explored boosting performance by combining GSURE with PnP denoisers.
Several visual examples of GSURE compared to DIP, with and without PnP denoiser, are presented in Fig.~\ref{fig:comparison_sr_scenario5}.

\section{Adapting a network to the input at inference time} 
\label{sec:test_time_adaptation}

The main limitation of ``pure" internal learning, where models are being trained from scratch based on $x_0$, is the potential performance drop due to not exploiting the massive amount of external data that is available for many tasks. 
Accordingly, in this section, we discuss different methods 
for adapting DNNs, which have been already pretrained offline, to better perform on the given test-time observations $x_0$. 
In Section~\ref{sec:test_time_adaptation_incorp} we discuss methods that incorporate offline external and test-time internal learning by using off-the-shelf pretrained models. 
In Section~\ref{sec:test_time_adaptation_meta} we discuss an alternative approach.
Instead of using off-the-shelf pretrained models, 
knowing in advance that a model is about to be fine-tuned at test time, allows using meta learning techniques in the offline phase with the goal of reducing the fine-tuning time at the inference phase. 
The methods that are discussed in this section are summarized in Table~\ref{table:internal_and_external}.

\subsection{Enhancing pretrained models via internal learning}
\label{sec:test_time_adaptation_incorp}

An immediate approach to incorporating external and internal learning is to use off-the-shelf pretrained models, which enjoy the existence of massive amounts of data and the generalization capabilities of deep learning, and fine-tune them at test-time using $x_0$ instead of training a DNN from scratch.
The end goal of this tuning is to specialize the network on the patterns of the specific unknown $x_{gt}$ to better reconstruct it. In practice, of course, the adaptation can use $x_0$ and the observation model $f$ rather than the unknown $x_{gt}$. (In the context of image editing the situation is slightly different, as will be discussed below.) 
The key risk, which is shared by all the methods in this section, is that exaggerated tuning will override or mask useful semantics/patterns that have been captured in the offline phase.
Thus, most of the methods below fine-tune pretrained DNNs using only a small-moderate number of iterations with relatively low learning rates.
As for the loss functions that can be used for fine-tuning, potentially, any data-fidelity term that is suitable for the observation model and does not depend on $x_{gt}$ can be used: least squares \eqref{eq:DIP_loss}, noise2void loss \eqref{eq:noise2void}, (G)SURE \eqref{eq:sure}, BP-term \eqref{eq:bp_term}, and various regularizations. Another possibility is to use loss functions that are based on extracting/synthesizing pairs of input-target patches from $x_0$ or utilizing the similarity of external images to $x_0$ to enlarge the fine-tuning data.
We will survey such approaches in this section.

Essentially, each of the methods discussed in Section~\ref{sec:internal_only} can be applied to a DNN that has been already pretrained, rather than with a DNN that is trained from scratch.
For example, in Section~\ref{sec:internal_only} we presented discussed using the SURE criterion for training a DNN for Gaussian denoising in a DIP-like manner: from scratch at test time using only the noisy $x_0$ (cf.~\eqref{eq:sure}).
Yet, utilizing SURE can be used also for offline training based on noisy external data with no clean ground truth samples \cite{soltanayev2018training}. Specifically, for each noisy training sample $x_{0,i}$ one uses \eqref{eq:sure} as the loss function instead of, e.g., an MSE criterion that requires having $x_{gt,i}$.
This pretrained denoiser can then be fine-tuned at test-time with $x_0$ using the same SURE criterion \cite{soltanayev2018training}.

Potentially, the direct fine-tuning approach, which is mentioned above, can be used for DNNs trained for general tasks (e.g., by using least squares loss or GSURE if $f$ is a linear operator).
However, it has been shown that a model that has been trained for a specific task, i.e., a specific instance of the forward/observation model $f$ and a specific class of images, suffers from a significant performance drop when it is tested on data that (even slightly) mismatch the training assumptions \cite{shocher2018zero,tirer2019super}. 
Therefore, a promising line of research focuses on fine-tuning of pretrained models that can be used to tackle solely the signal's prior, allowing for flexibility in the observation model at test time.

While generative models are natural candidates for models that can be used to impose only the signal's prior, the literature on PnP/RED \cite{venkatakrishnan2013plug,RED} has demonstrated that plain Gaussian denoisers can take this role as well.
Incorporating external and internal learning by fine-tuning pretrained CNN denoisers via the given $x_0$ and plugging them into a PnP/RED scheme was proposed in \cite{tirer2019super}.
This paper uses a PnP based on the proximal gradient optimization method (in the signal processing community this algorithm is oftentimes referred to as ISTA), which we present here.
The goal of this approach is to minimize w.r.t.~$x$ an objective function of the form
\begin{align}
\label{eq:data_plus_prior_objective}
\ell(x,x_0) + \beta s(x),
\end{align}
which is composed of a data-fidelity term $\ell(x;x_0)$ (e.g., the least squares term $\ell(x,x_0)=\|f(x)-x_0\|^2$), a signal prior term $s(x)$, and a positive hyperparameter $\beta$ that balances them. 
Traditionally, the prior term $s$ is a non-smooth explicit function. This motivates computing a gradient-step update only for the data term $\ell$, while handling $s$ with a proximal operation. Formally, starting with an initial $x^{(0)}$, the proximal gradient method reads as
\begin{align}
\label{eq:ista_step1}
\tilde{x}^{(t+1)} &=  {x}^{(t)} - \mu \nabla_x  \ell({x}^{(t)},x_0), \\
\label{eq:ista_step2}
{x}^{(t+1)} &=  \mathrm{prox}_{\mu \beta s(\cdot)}( \tilde{x}^{(t+1)} ),
\end{align}
where $\mu$ is a step size and
the operation
\begin{align}
\label{eq:def_prox}
\mathrm{prox}_{s(\cdot)}({x}) := \arg\min_z \,\, \frac{1}{2} \| {z} - {x} \|^2 + s({z})
\end{align}
is known as the proximal mapping of $s(\cdot)$ at the point ${x}$.
Notice that every iteration of the algorithm is decoupled into two steps: the first reflects the effect of the observation model and the second reflects the effect of the prior.
The core idea behind the PnP approach is recognizing \eqref{eq:def_prox} as the optimization problem that is associated with Gaussian denoising (which holds true even if $f$ reflects a different observation model), and thus instead of explicitly defining the function $s$ and computing \eqref{eq:ista_step2}, replacing \eqref{eq:ista_step2} with the execution of an off-the-shelf denoiser $h_{\sigma^*}(x)$ with a suitable noise level $\sigma^*$, namely, ${x}^{(t+1)} = h_{\sigma^*}(\tilde{x}^{(t+1)})$.

The PnP concept can be similarly utilized when minimizing \eqref{eq:data_plus_prior_objective} with other optimization methods that include a proximal mapping step like \eqref{eq:ista_step2}, e.g., ADMM \cite{venkatakrishnan2013plug}.
It has been shown that the approach is especially beneficial when using pretrained modern DNN denoisers, $h(\cdot;\theta)$ (where we omitted the denoiser's noise level for brevity), rather than model-based (e.g., sparsity-based) denoisers.


The work in \cite{tirer2019super} suggested adapting a pretrained CNN denoiser $h(\cdot;\theta)$ to $x_0$ before plugging it into a PnP scheme. Specifically, they use the scheme in \eqref{eq:ista_step1}-\eqref{eq:ista_step2}, but with a BP fidelity term \eqref{eq:bp_term} rather than a plain least squares term, and named the method IDBP-CNN-IA (in light of the similarity to the algorithm in \cite{tirer2018image}, which does not include image-adaptation (IA)).
In this case, \eqref{eq:ista_step1} with step-size $\eta=1$ forms a (back-)projection of $x^{(t)}$ onto the subspace $\{x: f(x) = x_0 \}$:
\begin{align}
\label{eq:bp_step}
\tilde{x}_{(t+1)} &= f^\dagger x_0 + (I - f^\dagger f) x^{(t)} \nonumber  \\
&= x^{(t)} + f^\dagger (x_0 - f x^{(t)}).
\end{align}
In many cases (e.g., deblurring and super-resolution), $f^\dagger$ (the pseudoinverse of $f$) can be implemented as efficiently as $f^T$.
In PnP/RED methods, the trainable model $h(\cdot;\theta)$ is a denoiser. Assuming that the given observation $x_0$ contains noticeable patterns of the original $x_{gt}$, the fine-tuning procedure can be done by 
synthetically injecting Gaussian noise into $x_0$. Specifically, the objective for the fine-tuning optimization is given by
\begin{align}
\label{eq:denoiser_ia}
    \sum_i \| h(P_i x_0 + \eta_i;\theta) - P_i x_0 \|_1,
\end{align}
where $P_i$ denotes patch extractions, $\eta_i$ is Gaussian noise (with standard deviation that matches the pretrained denoiser) that is randomly drawn at each optimization iteration, and the sum goes over the different patches (some of which are obtained by standard augmentations).
Essentially, the adapted denoiser scheme is akin to the ZSSR scheme in Fig.~\ref{fig:zssr_sketch}, but instead of synthetically downsampling $x_0$ it is synthetically noised.

IDBP-CNN-IA was examined in different super-resolution settings, where the assumption of the existence of patterns of $x_{gt}$ in $x_0$, possibly under some level of noise, is justified. The method was shown to outperform ZSSR (due to utilizing external data in the pretrained denoiser) as well as task-specific DNN super-resolvers in cases where $f$ and the noise level at test time mismatch those used in training (contrary to the flexibility of the PnP/RED approaches). 

As we mentioned above, it is also natural to specialize generative models on a given image $x_0$. Interestingly, while GANs are significantly different than denoisers, the recent generative approach based on score/diffusion models \cite{song2019generative,ho2020denoising} are based on offline training of Gaussian denoisers and iterative execution of them for image synthesis, in a way that shares similarity with PnP/RED methods. 
Accordingly, instead of fine-tuning a general purpose denoiser, \cite{abu2022adir} has suggested adapting off-the-shelf networks of diffusion models via a procedure that resemble \eqref{eq:denoiser_ia}.
Nevertheless, the proposed approach in \cite{abu2022adir}, dubbed ADIR, has a major difference from the one in \cite{tirer2019super}, which can essentially be applied also to any plain denoiser $h(\cdot;\theta)$ in PnP/RED frameworks.
Instead of building on the assumption that $x_0$ contains noticeable patterns of the original $x_{gt}$, which limits the applicability of the approach, they propose to use $x_0$ for retrieving related clean images from an external dataset, which will be used to tune the denoiser in lieu of $x_0$ in \eqref{eq:denoiser_ia}.
In more detail, it is proposed to look for K-nearest neighbors (kNNs) of $x_0$ in a diverse external dataset, with distance that is computed in neural embedding space---specifically, CLIP's embedding space. 
Empirically, it is shown that even for a degraded $x_0$ (e.g., a blurry version of $x_{gt}$) the kNNs of $x_0$ in CLIP's embedding space are similar to the unknown $x_0$, typically containing the same kind of key objects.

Since their invention \cite{goodfellow2014generative}, GANs have also been shown to be a powerful technique for generative modeling.
This has naturally led to using pre-trained GANs as priors in imaging inverse problems \cite{bora2017compressed}.
The outcome of training a GAN is a generator that maps a low-dimensional Gaussian vector $z \in \mathbb{R}^k$ to a signal space in $\mathbb{R}^n$ ($k \ll n$). To maintain notation consistency across the paper, we denote the generator by $h(\cdot;\theta)$, where $\theta$ are GAN's parameters that have been already optimized in the offline phase.
Consequently, given $x_0$, one can search for a reconstruction of $x_{gt}$ only in the range of the generator.
This can be done by setting $\hat{x}=h(\tilde{z};\theta)$, where $\tilde{z}$ is obtained by minimizing a data-fidelity term
\begin{align}
\label{eq:csgm}
\min_z \| f(h(z;\theta)) - x_0 \|^2.
\end{align}
This method, known as CSGM has been proposed in \cite{bora2017compressed}.
However, already in \cite{bora2017compressed}, and later with more focus in \cite{IAGAN}, it has been shown that while GANs can generate visually pleasing synthetic samples, the above procedure tends to fail to produce successful estimates of $x_{gt}$ as it is very unlikely that an image in the range of $h$ will sufficiently match an arbitrary image $x_{gt}$ (this issue has been oftentimes called ``limited representation capabilities" or ``mode collapse" of GANs). 

The image-adaptive GAN method (IAGAN) \cite{IAGAN} has suggested addressing this limitation via internal learning at test-time.
Specifically, the IAGAN approach suggests carefully tuning the generator's parameters and the latent vector simultaneously at test-time by
\begin{align}
\label{eq:iagan}
\min_{z,\theta} \| f(h(z;\theta)) - x_0 \|^2,
\end{align}
where $z$ is initialized by $\tilde{z}$  (optimizing $z$ alone in \eqref{eq:csgm}) and $\theta$ is initialized by the pretrained parameters.
Denoting the minimizers by $\hat{z}$ and $\hat{\theta}$, the latent image is estimated as $\hat{x}=h(\hat{z};\hat{\theta})$.
In the noiseless case, a post-processing back-projection step, as stated in \eqref{eq:bp_step} (with $\hat{x}$ in lieu of ${x}^{(t)}$), is suggested for boosting the results. 
IAGAN has different follow up works, 
such as DGP \cite{pan2020dgp}, which used it also for image editing, and PTI \cite{roich2022pivotal}, which applied it for image editing with StyleGAN specifically.

In Box 3 we present the visual results of IAGAN for compressed sensing and super-resolution and compare them with the results of DIP and CSGM. These results showcase the benefits of incorporating internal and external learning.

At this point, let us emphasize the differences and similarities between image reconstruction (solving inverse problems) and image editing in the context of fine-tuning a generative model.
The main difference is that in image editing the user is given the clean ground truth image, $x_0=x_{gt}$, which they aim to perceptually modify, e.g., change the object color or expression. 
Clearly, evaluating the success of such tasks is subjective in nature, and their implementation nowadays is typically based on pretrained generative models, such as GANs.
Since a clean image, $x_0=x_{gt}$, is given to the user, finding the best latent vector $z$ that expresses the projection of the image to the model's range (e.g., via \eqref{eq:csgm} with $f=I$) is easier than in inverse problems, where $x_0$ might be a seriously degraded version of $x_{gt}$.
Moreover, in editing, there is no risk of fitting noise/artifacts into the inversion.

Based on the above, one may ask: Why fine-tuning/internal-learning of a pretrained model is required for image editing?
The answer to this question is very similar to the reason fine-tuning is required for inverse problems \cite{IAGAN,roich2022pivotal}.
Even generative models that are specifically trained to allow easy editing, such as StyleGAN, tend to perform worse when they are given an arbitrary image rather than an image generated by the model itself.
The fine-tuning itself resembles \eqref{eq:iagan} (with $f=I$), but uses a metric like LPIPS (distance in some neural embedding) \cite{zhang2018perceptual} that is more aligned with human perception than MSE.
Also, even though there is no danger of fitting noise, the statement that we made at the beginning of this section still holds true: exaggerated tuning will override or mask useful semantics/patterns (required for editing, in this case) that have been captured in the offline phase.
Therefore, early stopping is still used also for image editing tasks \cite{pan2020dgp,roich2022pivotal}.

\begin{tcolorbox}[breakable,colback=blue!5!white,colframe=blue!75!black,title=Box 3: Adapting a pretrained GAN to the test image using IAGAN]

Instead of using internal learning for training DNNs from scratch using the observed image, it is beneficial to adapt DNNs, which have already been trained on massive external data, to the observed image. 

The prior information in GAN's generator, $h(\cdot;\theta)$, trained to map low-dimensional Gaussian vectors $z$ to data of the type of $x_{gt}$ (e.g., natural images of a certain class), can be utilized for estimating $x_{gt}$ from its observations $x_0=f(x_{gt})+e$, where $f$ is a known degradation model and $e$ is noise.

The popular CSGM method \cite{bora2017compressed} has suggested 
the estimator
$\hat{x}=h(\tilde{z};\theta)$, where $\theta$ is fixed to its pretrained values and $\tilde{z}$ is obtained by
$$
\min_z \| f(h(z;\theta)) - x_0 \|^2.
$$
Yet, CSGM fails to produce results that are aligned with the object in $x_{gt}$ due to the limited representation capabilities (``mode collapse") of generative models.

The image-adaptive GAN method (IAGAN) \cite{IAGAN} approach addresses this limitation via internal learning at test-time.
Specifically, IAGAN reconstructs the signal as $\hat{x}=h(\hat{z};\hat{\theta})$ where 
the generator's parameters $\hat{\theta}$ and the latent vector $\hat{z}$ are obtained by simultaneously minimizing
$$
\min_{z,\theta} \| f(h(z;\theta)) - x_0 \|^2.
$$
where $z$ is initialized by $\tilde{z}$ (optimizing $z$ alone) and $\theta$ is initialized by the pretrained parameters.

In the noiseless case, a post-processing back-projection step, as stated in \eqref{eq:bp_step} is suggested for boosting the results.

The figures below, taken from \cite{IAGAN} \tomt{and used by permission of the authors}, show the benefit of this test-time adaptation in several tasks. 

\vspace{3mm}

\includegraphics[width=0.24\linewidth]{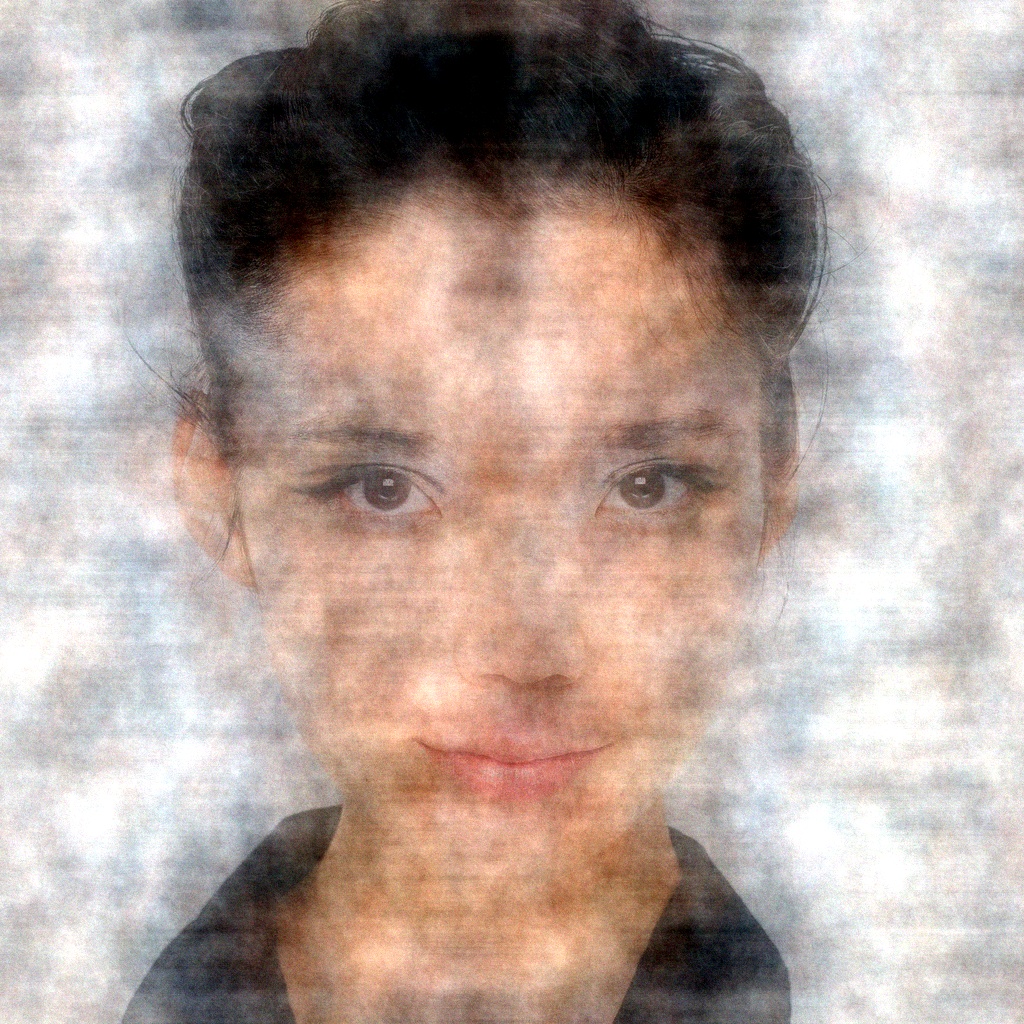}
\includegraphics[width=0.24\linewidth]{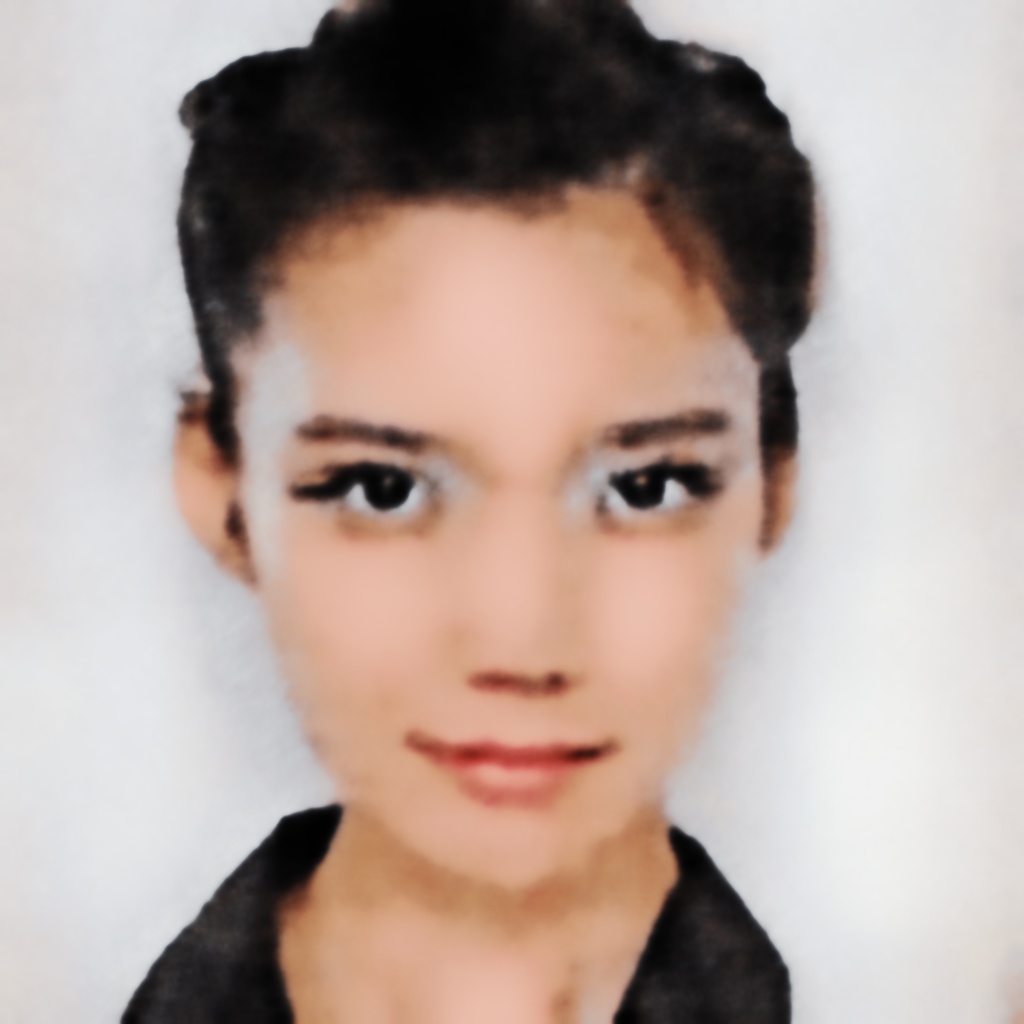}
\includegraphics[width=0.24\linewidth]{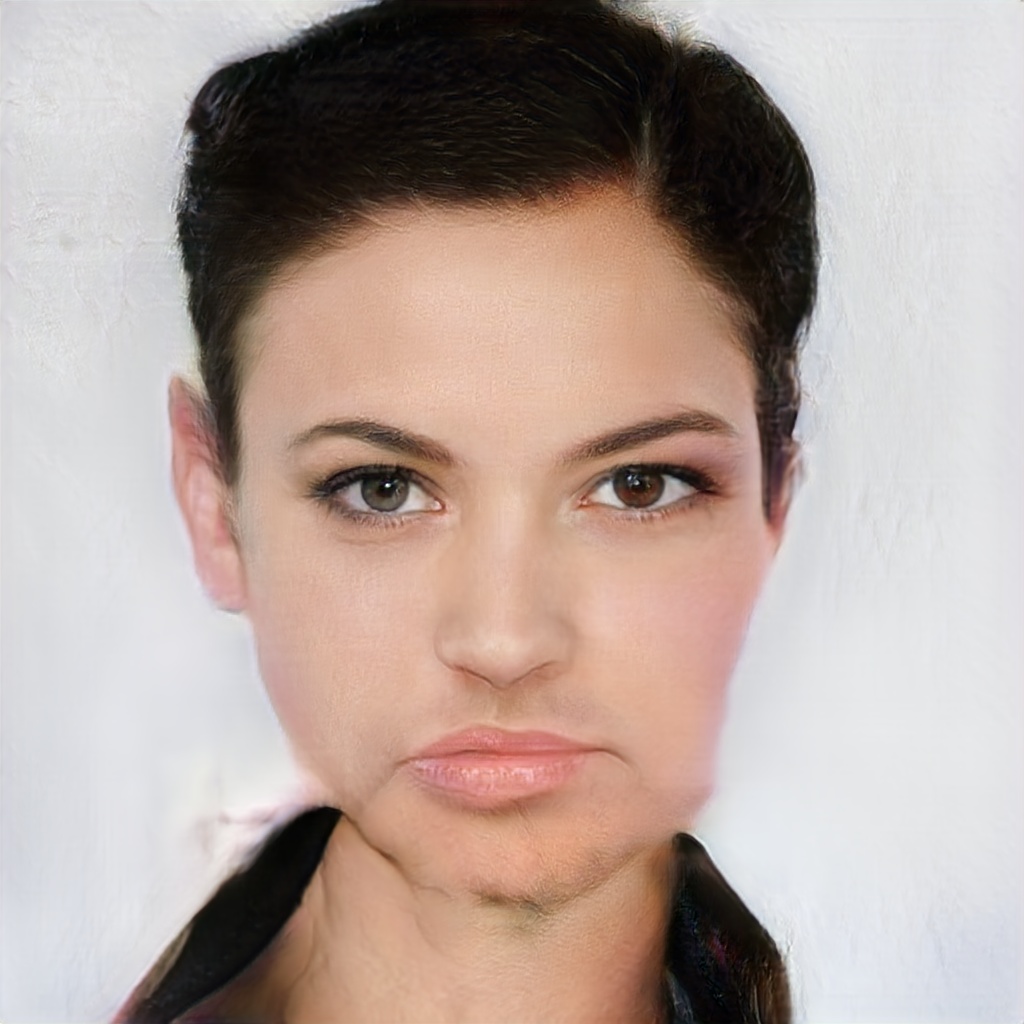}
\includegraphics[width=0.24\linewidth]{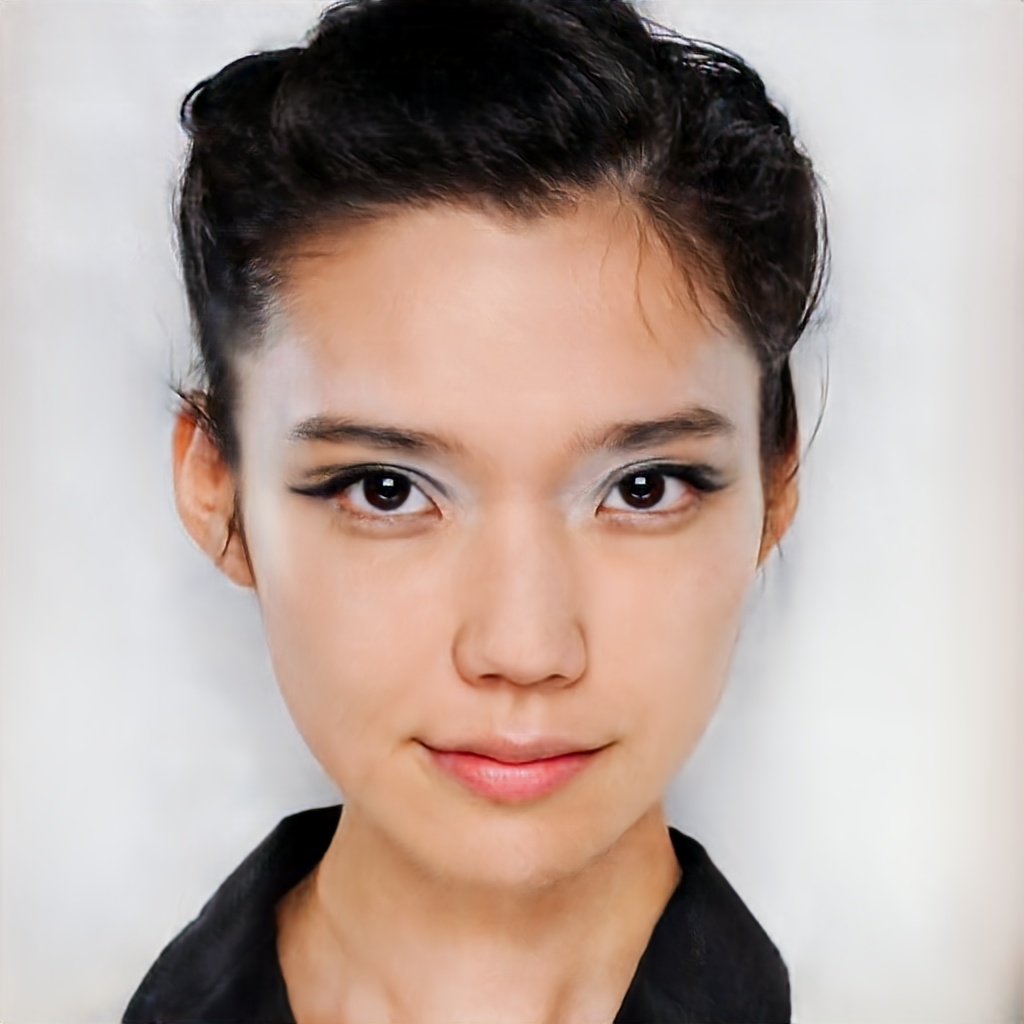}

\vspace{1mm}

\includegraphics[width=0.24\linewidth]{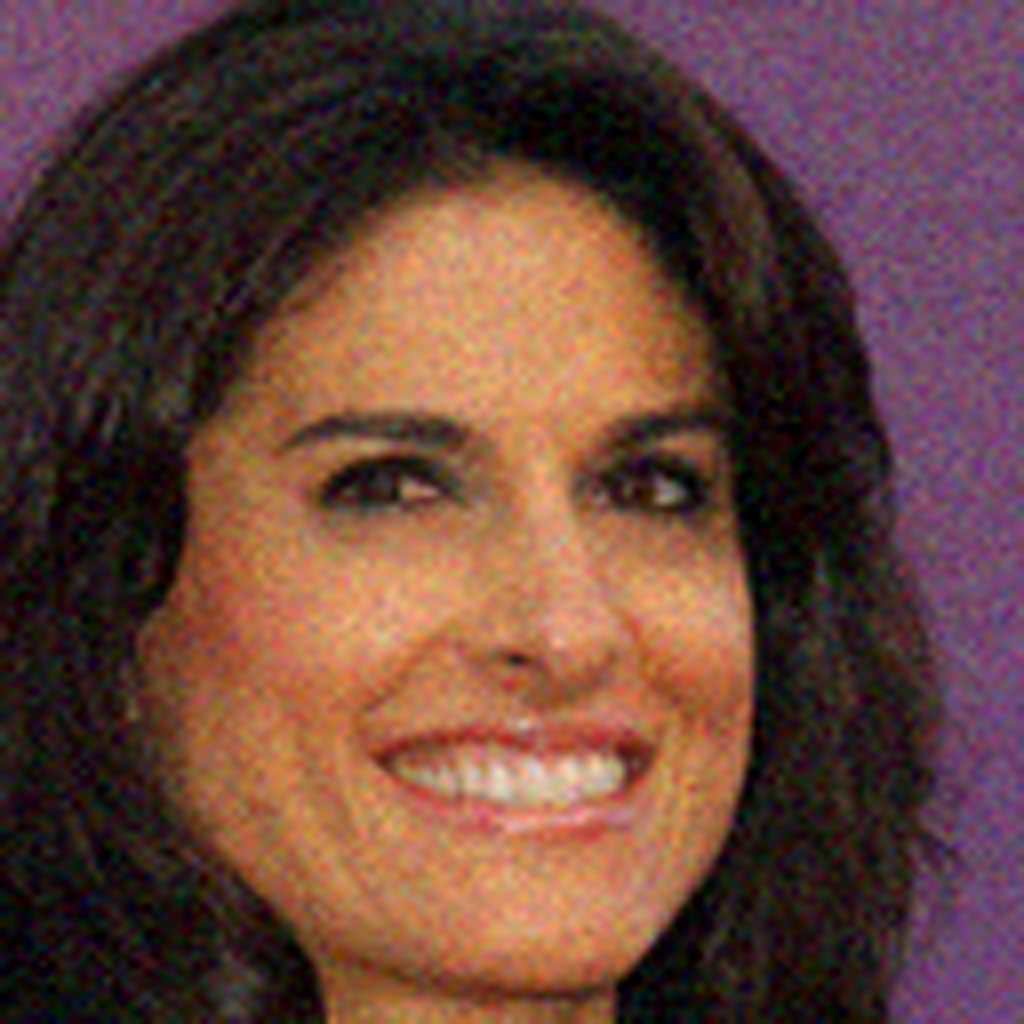}
\includegraphics[width=0.24\linewidth]{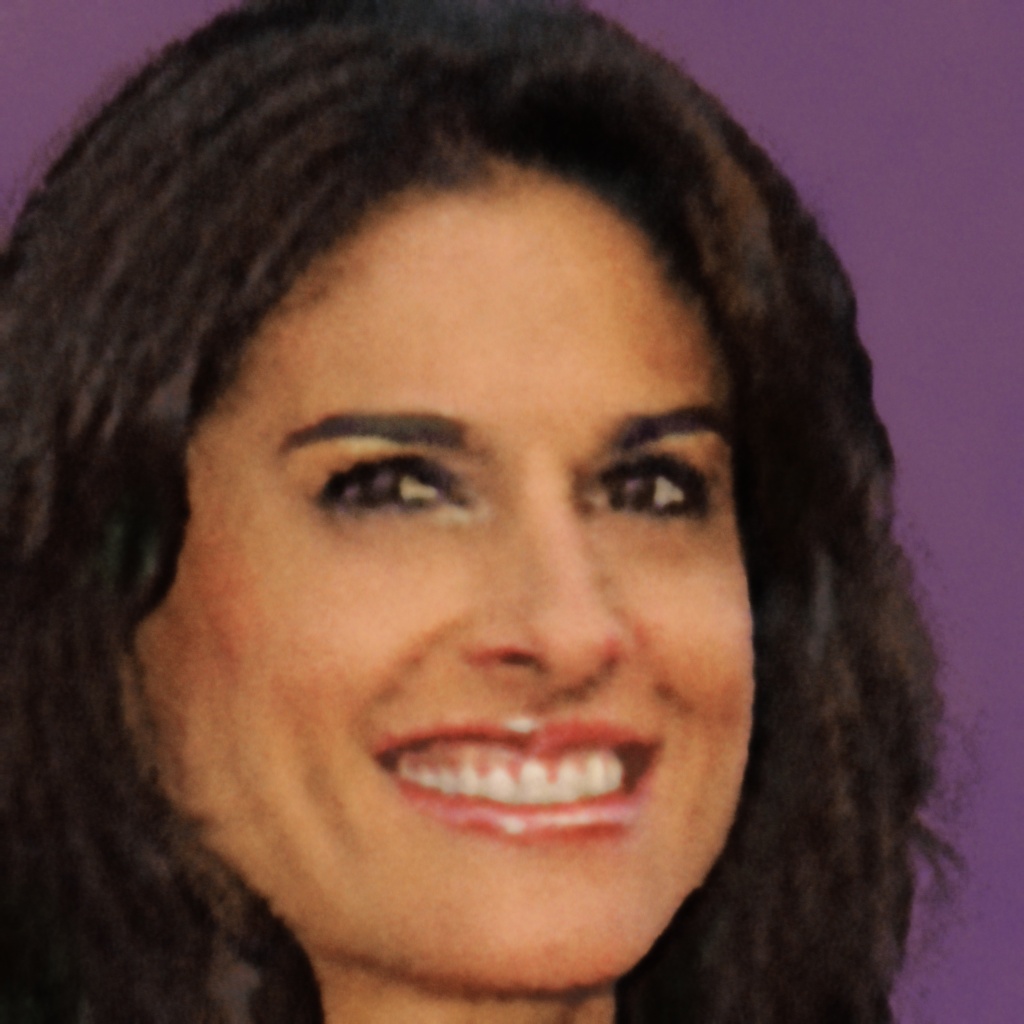}
\includegraphics[width=0.24\linewidth]{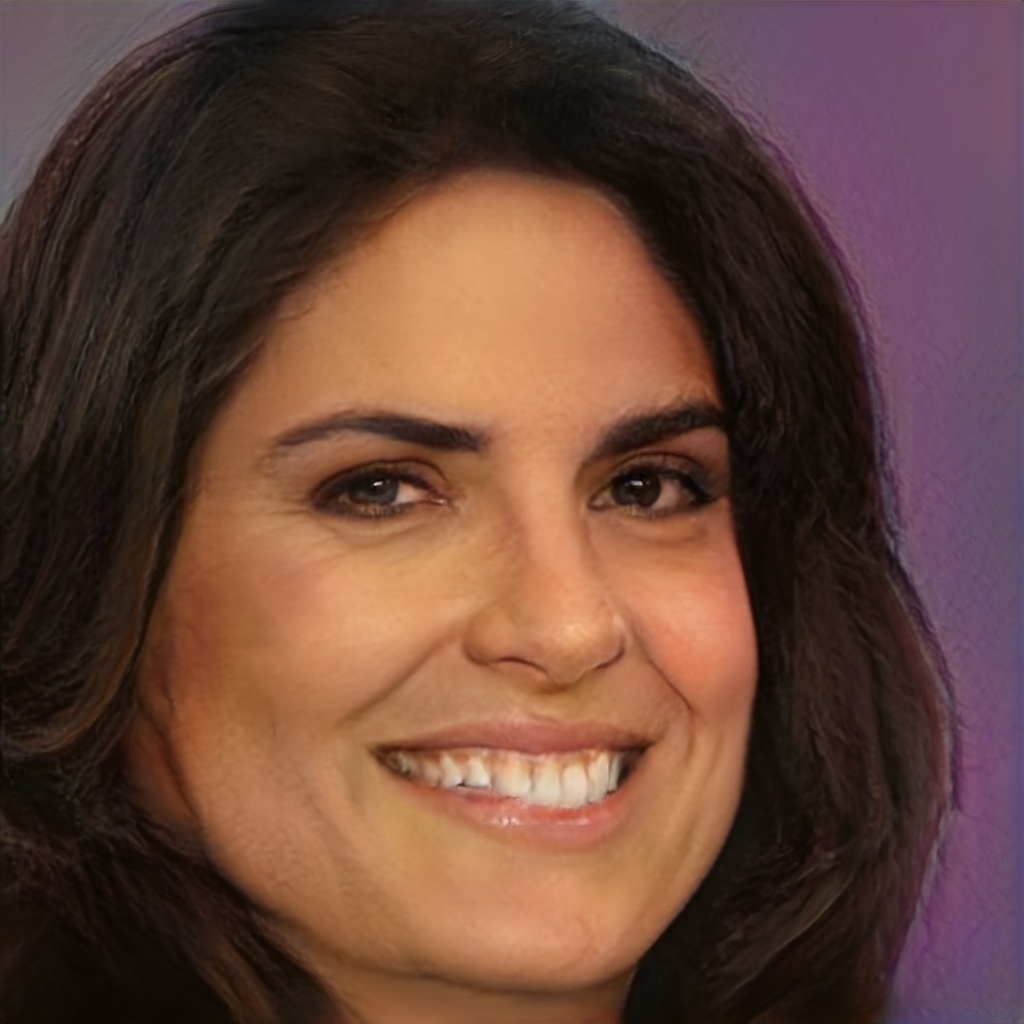}
\includegraphics[width=0.24\linewidth]{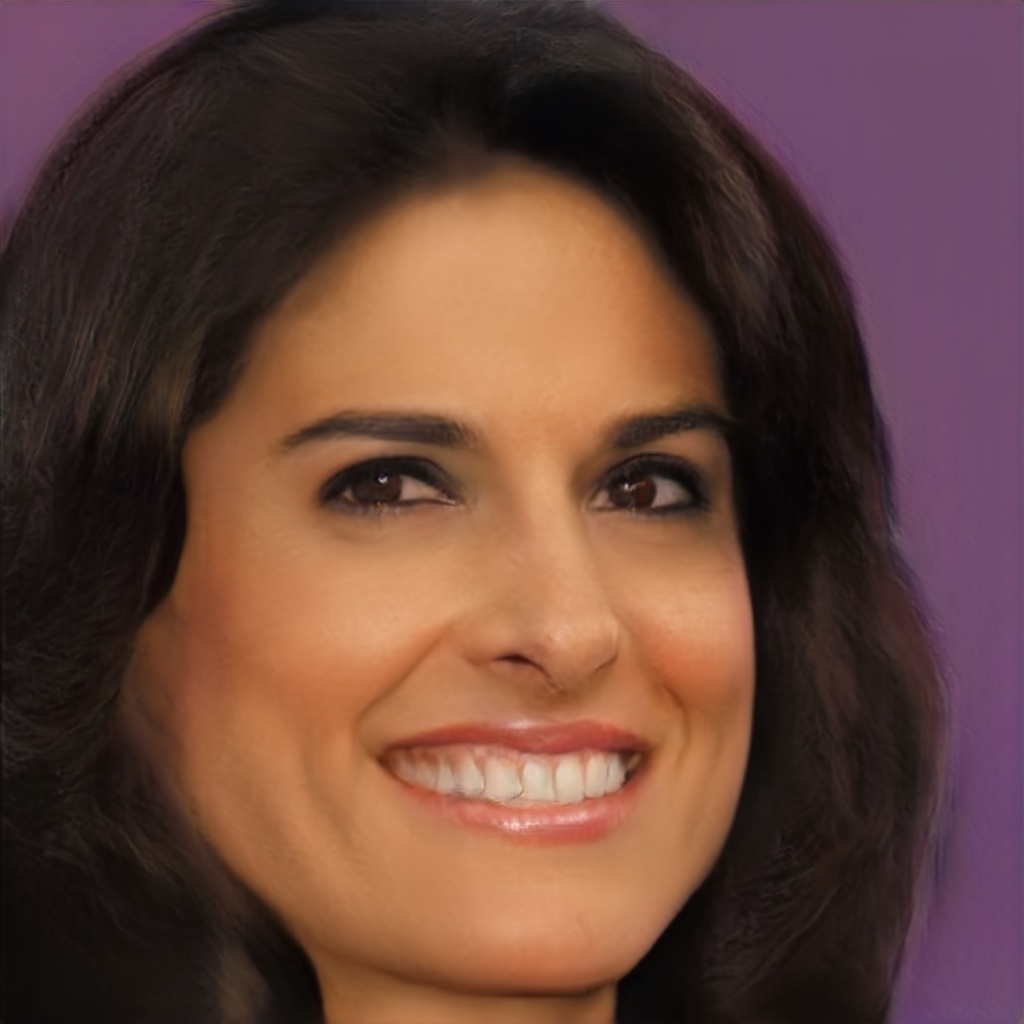}

Top: compressed sensing (CS) with 50\% pixels and a noise level of 10/255.
Bottom: super resolution (SR) x8 with bicubic kernel and noise level of 10/255.
From left to right: IFFT (CS) or bicubic upsampling (SR) of $x_0$, DIP (internal learning), CSGM (externally learned PGGAN), and IAGAN (internally+externally trained PGGAN).

\end{tcolorbox}

So far, we have discussed methods that fine-tune all the DNN's parameters.
Potentially, restricting the number of parameters being optimized in test-time can mitigate the need for early stopping.
The work in \cite{Daras2021ILO} considers using pretrained GANs for inverse problems and takes an approach similar to IAGAN \cite{IAGAN}: optimizing also the generator's parameters and not only the latent vector (cf. \eqref{eq:iagan} and \eqref{eq:csgm}).
However, they suggest optimizing only the intermediate layers of the network rather than all the parameters. As the focus of their paper is on expanding the range of the generator, 
robustness to overfitting noise has not been examined in \cite{Daras2021ILO}.
Focusing on the denoising task, the work in \cite{mohan2021adaptive} has suggested a GainTuning approach. Specifically, each learned filter (and its bias) in a pretrained CNN denoiser is multiplied by a newly introduced scalar gain parameter, which is initialized with 1.
Then, test-time fine-tuning involves optimizing only the gain parameters.
The authors examined the optimization objectives of both SURE and noise2void.
As typically CNN filters are of size $3 \times 3$ (so, each filter and its bias introduce 10 parameters), the number of parameters optimized in GainTuning is 10\% of the original model. This restricted optimization, which only affects filters' gains, has been empirically shown to resolve the problem of fitting noise.
Also, it has been shown that while in the synthetic experiments SURE has led to better results, when testing a denoiser, which is trained with simulations, on real electron-microscope images, noise2void tuning loss has been shown to be beneficial.
A limitation of restricting the tunable parameters may arise for out-of-distribution test images, which are significantly different from the training data.
Yet, improvements gained by the approach have been shown in \cite{mohan2021adaptive} for various out-of-distribution cases.

\subsection{Meta learning}
\label{sec:test_time_adaptation_meta}

In this subsection, we discuss methods that 
instead of using off-the-shelf pretrained models, use meta learning techniques in the offline phase with the goal of reducing the fine-tuning time at the inference phase \cite{soh2020meta,park2020fast}. 

Focusing on the super-resolution task (i.e., $f=d$ is a downsampling operator), MZSR \cite{soh2020meta} and MLSR \cite{park2020fast} proposed to incorporate ZSSR as the meta-test phase of the Model-Agnostic Meta-Learning (MAML) framework \cite{finn2017model}, which attempts to train a model such that it can be adapted to multiple tasks within a few optimization iterations at test-time.
Let us present MZSR (MLSR follows a similar idea).
The starting point is a relatively light DNN, $h(\cdot;\theta)$, similar to the one used by ZSSR, which (contrary to ZSSR) exploits offline external learning via a large scale dataset of ground truth images $\{ x_{gt,i} \}$, a bicubic downsampling kernel setting $\{ x_i=d(x_{gt,i}) \}$, and $\ell_1$-norm loss (cf. \eqref{eq:suptrain_}). 
Then, the DNN parameters are offline tuned by a meta-training scheme, which is applied after defining a family of Gaussian downsampling kernels $\{ d_j \}$ from which ``tasks" will be drawn.

An iteration of meta-training includes drawing an image (or mini-batch) from the dataset $x_{gt,i}$ and a task, i.e., a downsampling kernel $d_j$. 
Let the $j$th task loss be $\ell_j(\theta) = \|h(d_j(x_{gt,i});\theta)-x_{gt,i}\|_1$.
An update that will decrease this loss is given by
$$
\tilde{\theta}_j(\theta) = \theta - \alpha \nabla \ell_j(\theta).
$$
The objective that is actually being used for updating $\theta$ in this iteration attempts to minimize the total loss, averaged over all the tasks (kernels), at the look ahead point $\theta_j$:
\begin{align}
\min_\theta \sum_{j'} \ell_{j'} ( \tilde{\theta}_j(\theta) ).
\label{eq:maml}
\end{align}
Essentially, MAML tries to reach a point that is one gradient step far from being optimal for the sum of tasks.
Oftentimes, more than one gradient step is taken (unfolded) when defining $\tilde{\theta}_j(\theta)$ (e.g., \cite{soh2020meta} used 5 steps).
The main difference between \cite{park2020fast} and \cite{soh2020meta} is that the former uses meta-training based on ZSSR, without using $x_{gt,i}$, though it still has the same ``supervised" initial training.

At test-time, the meta-test fine-tuning is performed exactly as described for ZSSR in \eqref{eq:zssr}.
However, now instead of training the network from scratch the weights are initialized at the point obtained via MAML.
The gain of its offline training is that this approach has been shown to be competitive with ZSSR with only a single optimization step at test-time and outperforming it with more steps.

\begin{figure}
    \centering
  \includegraphics[width=250pt]{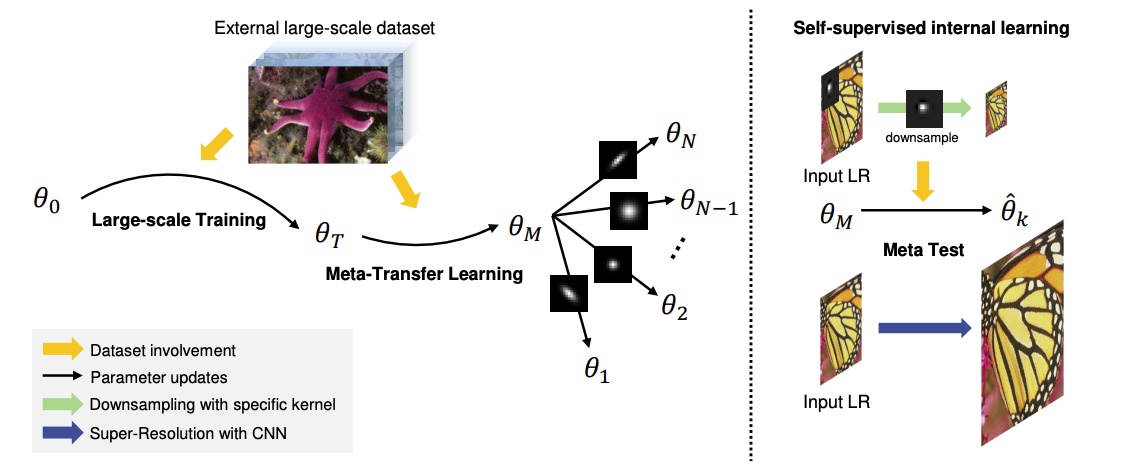} 
  \caption{``Meta-transfer learning for Zero-shot Super-Resolution" (MZSR) approach. During meta-transfer learning, the external dataset is used. During 
internal learning is done during meta-test time. From random initial point $\theta_0$, large-scale dataset DIV2K with bicubic
degradation is exploited to obtain $\theta_T$. Then, meta-transfer learning learns a good representation $\theta_M$ for super-resolution tasks with diverse blur kernel scenarios. 
In the meta-test phase, ZSSR-based self-supervision is used, with the test blur kernel and the test image. 
Figure taken from \cite{soh2020meta} \tomt{and used by permission of the authors}.}
\label{fig:MZSR_sketch}
\end{figure}

The MZSR scheme is presented in Fig.~\ref{fig:MZSR_sketch}. 
Though being conceptually elegant, note that the main difficulty of this approach is obtaining a successful meta-learning stage, as MAML is known to suffer from stability issues.

\section{Open problems and challenges} 


We conclude this review with a discussion of open problems and challenges in internal learning and the role of signal processing in addressing them.

As described in this paper, the main limitation of ``pure" internal learning, where models are trained from scratch based on $x_0$, is the potential performance drop due to not exploiting the massive amount of external data that is available in many tasks. On the other hand, the main benefit is bypassing any assumptions made in the offline training phase (e.g., on the ground truth data and the observation model) that can mismatch the situation at test-time.

Therefore, whenever informative ground truth external data is available, a good balance between the pros and cons of internal and external learning can be obtained by offline training powerful models that will serve only as the signal's prior. Such models may be generative models (e.g., GANs and diffusion models) or, as discussed, even plain denoisers. Another alternative is using deep unfolding to generate neural architectures with their optimization objectives \cite{Monga2021Algorithm,Shlezinger2023Model}. 
At test-time, these models may be adapted to observations and used for restoring the latent test image.
The main limitations of this adaptation are the dependency on the level of degradation in the observations and the additional computational cost at test-time.

Tools and ideas from signal processing can be utilized to mitigate these issues. For example, since the ``training data" for internal learning is merely the degraded image, one can use traditional signal processing methods to enhance the observations. Furthermore, the data that is used at test-time may be enriched via transformations of observations beyond regular augmentations.
As for reducing the fine-tuning time, one may try to utilize ``universal representations", such as wavelet for images, within the network in lieu of learnable parameters that may overfit in the offline training phase and by that reduce the number of parameters that need to be fine-tuned.
Alternatively, low-rank matrix factorization strategies can be utilized, as recently done in \cite{hu2021lora}, in order to reduce the dimension of the optimization variables at test-time.

Another challenge is the ``blind setting", where the observation model $f$ in \eqref{eq:observation_model} is fully or semi-unknown.
The focus of this paper is on non-blind cases, where $f$ is known. In fact, in the blind setting, external learning is oftentimes not possible, while some of the methods discussed here can be applied after an initial phase of estimating $f$ (e.g., kernel estimation in deblurring and super-resolution). The quality of this estimation obviously affects the succeeding image reconstruction.
In the signal processing community there are various approaches to estimating such nuisance parameters. Incorporating them with internal learning may boost their performance, as shown in \cite{bell2019blind,Hussein20Correction}.

Finally, note that theoretical understanding of internal learning is at its infancy. 
For example, no recovery guarantees, such as those that have been established in the compressed sensing literature, exist when the signal is parameterized by a neural network. Theoretical advances of this kind would be highly significant.


\section*{Acknowledgment}
TT was supported by the Israel Science Foundation (ISF) grant No.~1940/23. 
RG was supported by the ERC-StG grant No.~757497, by the Tel Aviv University Center for AI and Data Science, and by a KLA grant. 
SYC was supported by Institute of Information \& communications Technology Planning \& Evaluation (IITP) grant funded by the Korea government(MSIT) [NO.2021-0-01343, Artificial Intelligence Graduate School Program (Seoul National University)] and the National Research Foundation of Korea (NRF) grant funded by the Korea government (MSIT) (No. NRF-2022R1A4A1030579, NRF-2022M3C1A309202211).
YE was supported by the European Research Council (ERC) under the European Union’s Horizon 2020 research and innovation program grant No. 101000967, by the Israel Science Foundation grant No. 536/22, and by an Amazon research fund.

\bibliographystyle{IEEEtran}
\bibliography{main}

\end{document}